\documentclass[journal]{vgtc}                
\pdfoutput=1
\ifpdf
  \pdfoutput=1\relax                   
  \usepackage{graphicx}                
  \DeclareGraphicsExtensions{.pdf,.png,.jpg,.jpeg} 
\else
  \usepackage{graphicx}                
  \DeclareGraphicsExtensions{.eps}     
\fi%

\graphicspath{{figures/}{pictures/}{images/}{./}} 
\usepackage{paralist}
\usepackage{algorithm} 
\usepackage{algorithmicx}  
\usepackage{algpseudocode}  
\usepackage{amsmath}
\usepackage{color,soul}
\usepackage{capt-of}
\usepackage{microtype}                 
\PassOptionsToPackage{warn}{textcomp}  
\usepackage{textcomp}                  
\usepackage{mathptmx}                  
\usepackage{times}                     
\usepackage{cite}                      
\usepackage{tabu}                      
\usepackage{booktabs}                  
\usepackage{bbding}
\usepackage{pifont}
\usepackage{wasysym}
\usepackage{amssymb}
\definecolor{ColorName}{RGB}{0,0,0}  
\definecolor{ColorNameTODO}{RGB}{0,0,0}  
\definecolor{ColorVISRevise}{RGB}{0,0,0}  

\vgtccategory{Research}

\title{VAC$^2$: Visual Analysis of Combined Causality in Event Sequences }

\author{  
Sujia Zhu, Yue Shen, Zihao Zhu, Wang Xia, Baofeng Chang, Ronghua Liang, Guodao Sun*
}
\authorfooter{
E-mail: godoor.sun@gmail.com
}

\abstract{
Identifying causality behind complex systems plays a significant role in different domains, such as decision making, policy implementations, and management recommendations. 
However, existing causality studies on temporal event sequences data mainly focus on individual causal discovery, which is incapable of exploiting combined causality. 
To fill the absence of combined causes discovery on temporal event sequence data, 
eliminating and recruiting principles are defined to balance the effectiveness and controllability on cause combinations. 
We also leverage the Granger causality algorithm based on the reactive point processes to describe impelling or inhibiting behavior patterns among entities. 
In addition, 
we design an informative and aesthetic visual metaphor of "electrocircuit" to encode aggregated causality for ensuring our causality visualization is non-overlapping and non-intersecting. 
Diverse sorting strategies and aggregation layout are also embedded into our parallel-based, directed and weighted hypergraph for illustrating combined causality.
Our developed combined causality visual analysis system can help users effectively explore combined causes as well as an individual cause.  
This interactive system supports multi-level causality exploration with diverse ordering strategies and a focus and context technique to help users obtain different levels of information abstraction.
The usefulness and effectiveness of the system are further evaluated by conducting a pilot user study and two case studies on event sequence data.
} 


\CCScatlist{ 
 \CCScat{K.6.1}{Management of Computing and Information Systems}%
{Project and People Management}{Life Cycle};
 \CCScat{K.7.m}{The Computing Profession}{Miscellaneous}{Ethics}
}

\teaser{
  \centering
  \includegraphics[width=\linewidth]{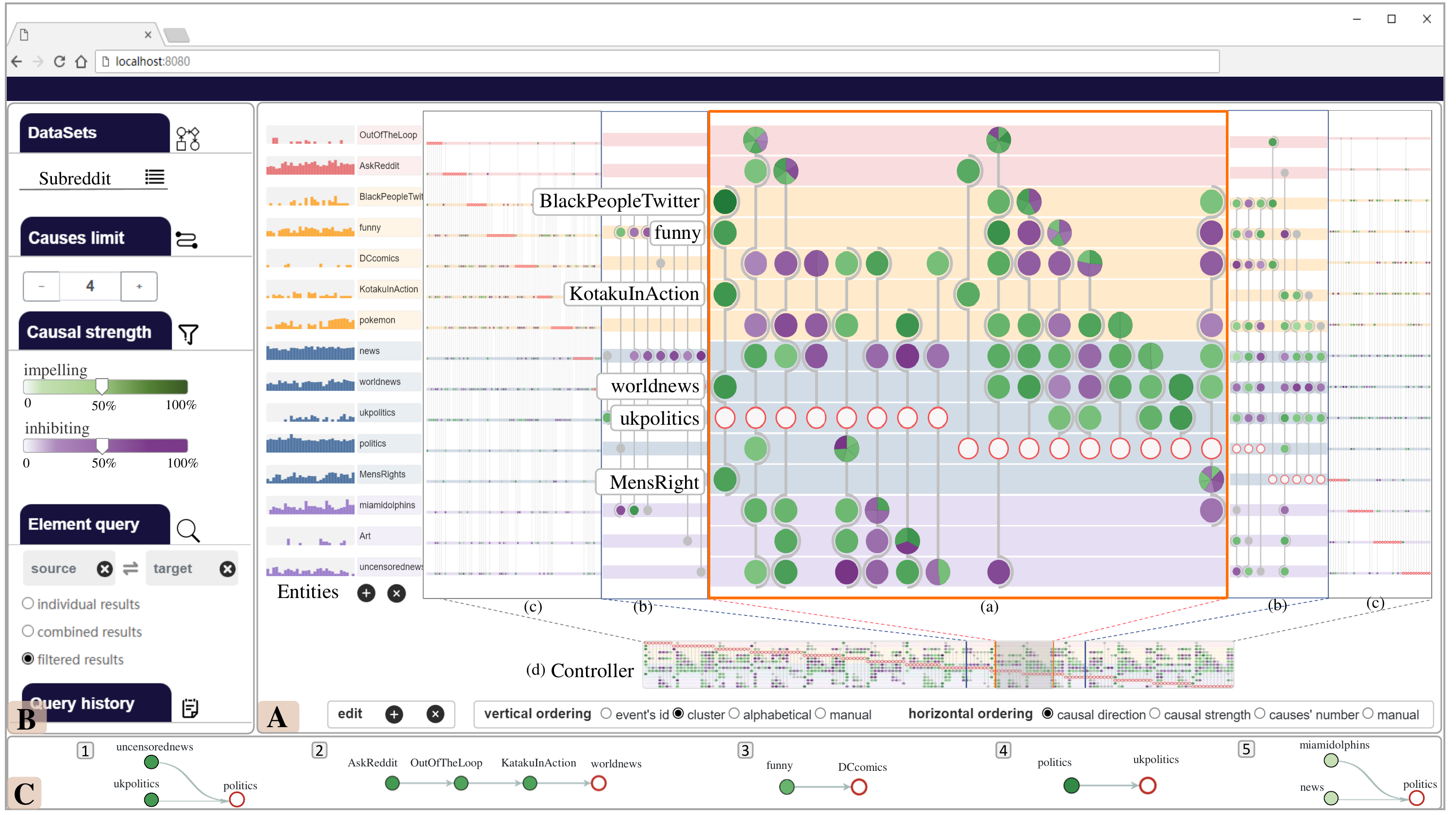}
  \vspace{-0.6cm}
  \captionof{figure}{\textit{User interface}:  (A) Combined causality panel for illustrating the individual and combined causes at different level of causality abstraction with controller and diverse ordering. 
  (B) Parameters configuration panel for providing users filtering causality.
  (C) Propagation panel for finding and tracking causal propagation and storing the history of causal propagation.
  }
  \label{Figure:overview}
  \vspace{-0.1cm}
}

\vgtcinsertpkg

\begin{document}



\maketitle

\section{Introduction}



Causal discovery mainly focuses on revealing deep and strong correlations among entities. 
Human activities form a large number of event sequences, where potential causal relations exist among these entity events.
Discovering causalities is a ubiquitous and crucial task in many applications such as medical treatment, marketing research, products recommendation, and policy implementation.

%

In recent years,
researchers have been devoting efforts to analyze causal effects~\cite{angrist1996identification} and causality problems~\cite{guo2020survey}.
Most causality researchers focus on the individual causality discovery and pay little attention to combined causality discovery.
{\color{ColorVISRevise}
Multifaceted factors are deemed/considered as the a combined cause.
An scenario/effect are commonly caused by confounding mutual impacts.
For example, High temperature, oxygen, and wood together result in fire,
and none of them can cause fire individually.
Individual causality discovery obtain result that high temperature, oxygen, and wood be the causes, respectively.
Because these researches are conducted under a assumption of neglecting all the other necessary factors.
}
Existing approaches~\cite{jin2020visual} on combined causes discovery can only be applied to multi-dimensional datasets 
with constraint-based algorithms.
No existing algorithms are proposed to discover and analyze combined causal factors for temporal event sequence dataset.
In addition,
little attention are attracted to detailed description of causality, such as impelling or inhibiting behavior
{\color{ColorVISRevise} that increase or decrease the occurrence possibility of a certain event}.

Meanwhile,
many researchers on causality visualization have devoted to simple directed graph layout for efficiently and intuitively analyzing causality in certain scenarios, such as group clustering, outlier finding, and topology detecting.
Visual illusion and visual clutter always exist because an entity may be contained in various combined causes or effects, and elements in combined causes may be mistaken for several separate causes.
Thus,
effective, intuitive and formative visualizations are desired to display complex multiple combined causality from overview level and individual causality at instance level for the event sequence dataset. 

To analyze combined causality as well as individual causality for the event sequence dataset, three challenges remain to be addressed. 
Causes found or observed in the causality discovery may be mutually affected by other existed causes due to potential impelling or inhibiting behaviors.
Thus,
the first one is to reveal impelling or inhibiting behavior of causality.
Traditional association rules mining techniques can neither certainly demonstrate causality nor the impelling or inhibiting behaviors of causality.
Although certain approaches are designed for causality discovery, 
they can not describe and explain the impelling or inhibiting behaviors.
Causality results of these approaches tend to be ambiguous, uncertain, and spurious.
Complex interactions exist in a large number of activities, making causal discovery a particularly challenging problem.
The second one is the combined causal discovery of entities from the temporal event sequence dataset.
The exploit number of combined causes result in the computational complexity.
Thus, eliminating redundant combined causes and reducing the number of valid combined causes are significant tasks for combined causes discovery.
In addition,
events may mutually resist or promote each other, resulting indirect, redundant, and spurious causality easily appearing in a cause combination.
Therefore,
eliminating the spurious causality and identifying useful combined causes serve as challenging for the event sequence dataset. 
The third one is visual clutter and visual illusion of complex combined causality visualization.
Our proposed directed and combined causal relations detected from our causal discovery algorithm form a hypergraph, where certain entity may be contained in several causes sets and elements in these causes sets may have causal relations.
Thus, visual clutter and visual illusion always occur as a result of excessive edge crossings and edge combinations.
Causality is commonly visualized with a simple directed graph, by which points and lines with arrows represent entities and directions of causality among entities, respectively.
This type of traditional graph drawing can only show a kind of causal relationship between a single pair of cause and effect. 
Individual cause and combined causes can not be distinguished, resulting in visual confusion and illusion, which is challenging for illustrating combined causality.


To address the first challenge, 
we leverage Granger causality algorithm, which is regarded as the basis of causal discovery for temporal data and is defined regarding the predictability and the temporal ordering of events. 
This causal discovery algorithm is based on the Reactive point process~\cite{ertekin2015reactive}, which unveils impelling and inhibiting behavior patterns to find both individual and combined causality.
To address the second challenge,
filtering rules are defined for candidate cause combinations by abandoning redundant combination entries and recruiting the useful combination entries.
To deal with the third challenge, 
we design an informative and aesthetic visual metaphor of "electrocircuit" to encode and aggregate ``AND" causes and ``OR" causes for ensuring our causality visualization non-overlapping and non-intersecting. 
We also embed diverse sorting strategies, aggregation layout, a focus and context approach, and smooth interaction techniques into our parallel-based and directed hypergraph for illustrating combined causality.
In addition,
we design a combined causality analysis system to help users explore combined causal factors for event sequence dataset.
The system consists of three panels. 
In configuration panel, users can set parameters to filter causality.
In the combined causality panel, a parallel-based directed hypergraph is designed for analyzing causality with different levels of information abstraction of impelling and inhibiting behaviors.
In addition,
the characteristics of entities used in a statistic component can help users further understand the causal relations among entities.
In propagation panel, two arbitrary entities are tracked to illustrate the influence propagation.  VAC$^2$ is provided as an open source project\footnote{https://zhuzihao-hz.github.io/VAC2/}.

In addition, 
we demonstrate the usefulness and effectiveness of our methods by conducting two case studies 
and a pilot user study on visual design and task analysis. 

To summarize, our key contributions include:

\begin{compactitem}
	\setlength{\itemsep}{1pt}
	\setlength{\parsep}{1pt}
	\setlength{\parskip}{1pt}

	\item
	We develop an individual causal discovery algorithm, which reveals impelling and inhibiting behaviors among entities for the event sequence datasets. 
	
	\item 
	We propose a novel combined causality visualization, which incorporates a causal discovery mechanism to detect combined causes among entities. 

	\item 
	We design a causality analysis system that aims to interpret combined and individual causal relationships for efficient exploration.



\end{compactitem}

\vspace{-0.3cm}
\section{Background and Related Work}
Our work aims to visually analyze both individual and combined causes of events in the event sequence dataset.
Thus,
the related works are concerned about causal discovery and visual analysis of causality. 

\vspace{-0.2cm}
\subsection{Causal discovery}


Causality uncovers deep relationships between the causes and effects that we can intuitively perceive. Such relationships are difficult to accurately define compared with traditional association relationships.
Fortunately,
There are fundamental researchers~\cite{angrist1996identification} and comprehensive reviews~\cite{guo2020survey} on causality analysis.
In this paper, we subsume the causal discovery algorithm into two types, namely, individual cause discovery and combined causes discovery

For individual cause discovery~\cite{angrist1996identification}, existing approaches can be divided into constraint-based, score-based, and hybrid algorithms.
Constraint-based approaches, including PC~\cite{colombo2014order}, SGS~\cite{spirtes1991probability}, FCI~\cite{spirtes2000causation}, and their  extended algorithms, mainly implemented exponential numbers of conditional independence (CI) tests to delete the edges of independent variables in the hypothetical and fully connected graph.
However, these approaches can not quantify the causal strength of the variables.
Score-based approaches, including BIC score, GES~\cite{chickering2002optimal}, and F-GES~\cite{ramsey2017million}, introduced structural equations or kernel functions to determine/quantify the causal strength among causes and effects.
However, these approaches commonly involve a large number of parameters, which have significant impacts on causal results during the optimization process. 
Hybrid approaches, including ARGES~\cite{nandy2018high} and MMHC~\cite{tsamardinos2006max}, combine constraint-based and score-based approaches by their consistency results to achieve causal discovery. 
However, when two results obtained by constraint-based and score-based approaches are in conflict, making a decision poses a challenge.
Ground truth of causal relationships is possibly lost via utilizing the hybrid approaches.

This paper aims to analyze the causal relations for temporal datasets,
thus, we focus on algorithm or technique advances~\cite{mirza2014analysis} on causal discovery for temporal event sequence data.
Granger causality test~\cite{seth2007granger} is the most commonly used causal inference in the temporal dataset. 
The intervention concept is introduced to determine the causal relations via judging whether a certain variable changed after another's intervention.
Jin et al.~\cite{jin2020visual} proposed a user-feedback mechanism to enhance the performance of automatic causality analysis by utilizing Granger causality on the Hawkes processes~\cite{xu2016learning} for the event sequence datasets.
However, the above approaches on individual causal factor discovery can not uncover impelling and inhibiting behaviors of causality among variables or entities.

For combined causes discovery, existing causality studies mostly focus on individual cause discovery, and little attention is paid to combined causes discovery~\cite{zhang2021combined}.
Nevertheless, 
combined causes can be identified by performing kinds of individual cause discovery, such as traditional CI test~\cite{ma2016mining}.
Combined causes discovery poses a challenge of low efficiency because many researchers proposed approaches to improve the efficiency of combined causes discovery, such as a novel model~\cite{zhang2021combined}, an efficient algorithm~\cite{jin2012discovery}~\cite{li2013mining}~\cite{ma2016mining}, and a frequent itemset mining algorithms~\cite{alharbi2015conjunctive}.
First, combined causes rules and principles~\cite{jin2012discovery} are defined via leveraging association and partial association based on the idea of causal relationships being persistent.
Then, many researchers have proposed kinds of approaches to trade-off complexity of computation and accuracy. 
An additive noise model~\cite{zhang2021combined} is introduced to subsume all possible combined causes into three categories, give them formal and complete definitions, and detect combined causes.
A multi-level approach~\cite{li2013mining} is proposed to efficiently identify and qualify potential combined causes by utilizing an efficient association mining approach in observational data.
Ma et al.~\cite{ma2016mining} also proposed a multi-level approach to detect the combined causes via HITON-PC algorithm, an efficient and commonly used local causal discovery method by CI test, when given a target variable. 
CCCRUD~\cite{alharbi2015conjunctive} and DCCRUD~\cite{alharbi2015disjunctive} are proposed by employing frequent itemset mining algorithms, to identify disjunctive combined causal rules and conjunctive combined causal rules from uncertain data.

Our work aims to analyze the individual and combined causes among entities.
However, these approaches on combined causes discovery are designed for multiple dimensional static datasets, but can not be applied to temporal event sequence datasets.



\begin{figure*}[htb]
	\centering
	\includegraphics[width=1.01\linewidth]{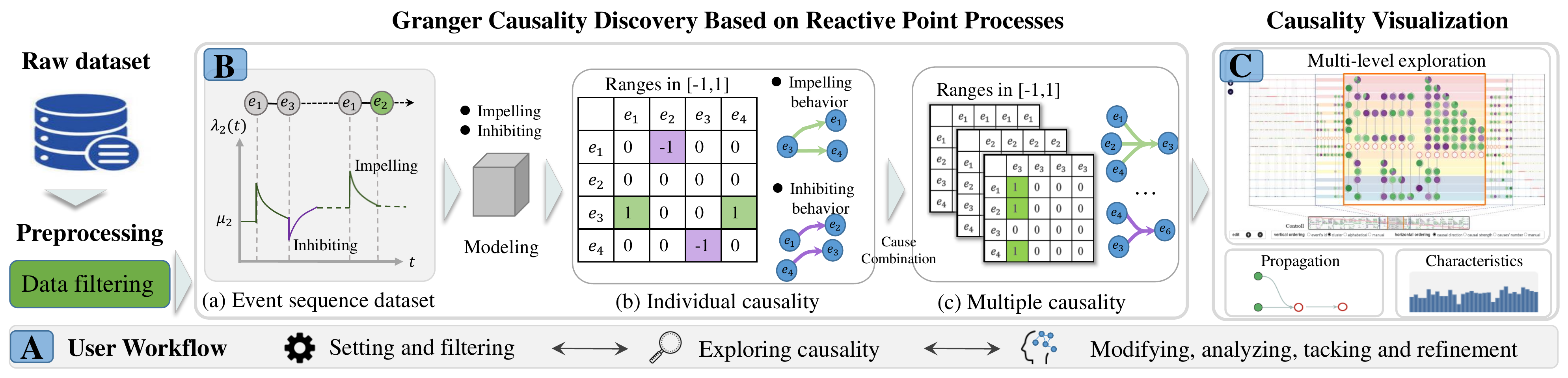}
	\vspace{-0.7cm}
	\caption{ 
The system architecture (A) and the pipeline of our combined causal discovery algorithm (B) which impelling and inhibiting reveals behaviors. 
 The impelling and inhibiting behaviors discovery is based on granger causality, which consists of three key steps: (a)training a reactive point processes model to fit event sequence dataset, (b)discovering impelling or inhibiting behaviors on causality, 
 which ranges in [-1, 1] between two entity, and (c)inferring the cause combinations based on the individual causal discovery.
 The framework of our individual and combined causality analysis system (C) can be divided into the following steps: preprocessing and filtering raw dataset, modeling combined causal discovery and exploring directed hypergraph with diverse ordering operations and interaction techniques. 
	}
	\label{Figure:pipline}
	\vspace{-0.5cm}
\end{figure*}

\subsection{ Visual Causality Analysis }
%
The basic visual analysis of researchers for user perception and cognition on causality exist~\cite{xiong2019illusion}~\cite{yen2019exploratory}. 
They contribute experiments~\cite{kale2021causal} demonstrating an approach to provide design guidelines with visual encodings~\cite{cleveland1984graphical}.


Causal relations are commonly illustrated with a directed acyclic graph, where a node serves for a type of entities and a link with an arrow encodes the causal relation and direction between two kinds of entities, such as 
Causalflow~\cite{xie2020causalflow} and SeqCausal~\cite{jin2020visual} 
for event sequence dataset, 
as well as 
Compass~\cite{deng2021compass}, Causality Explorer~\cite{xie2020visual} and the Visual Causality Analyst~\cite{wang2015visual} for multiple dimensional datasets.
The above studies are designed for visual exploration of simple directed graph.
In our proposed combined causality analysis, entities can be reached only after all causes in a combined combination have been implemented while this condition is lost in the traditional graph.
Thus, these kinds of traditional graph drawings can only show a causal relationship between a single pair of cause and effect, which can not be intuitively distinguished with the combined causes.



A hypergraph~\cite{alsallakh2016state}~\cite{fischer2021towards} is a generalization of a standard one-to-one-relationship graph.
Hypergraphs is usually visualized as standard node-link representations, subset-based hypergraph, matrix-based hypergraph and parallel-based hypergraph.
Many surveys~\cite{ouvrard2020hypergraphs}~\cite{vehlow2015state} and researchers~\cite{kaufmann2008subdivision}~\cite{qu2021automatic} on hypergraph visualization exist.
Hyperedges denote entity relations, and each edge set shares the same source entities or target entities.
Researchers utilize hypergraphs to effectively uncover correlation and the correlation patterns among entities in certain applications, such as co-authorship networks~\cite{kumar2015co},
communication~\cite{ren2020weighted}, security~\cite{wang2020security}, and economy~\cite{ranshous2017exchange}.
Meanwhile, three kinds of illustrations~\cite{disixSixmethods} are introduced for the transformation of layered hypergraphs, and these illustrations have their own advantages in the performance of simplicity, efficiency, and consistency in processing times.
Standard node-link hypergraph drawing is the most common approach.
Kinds of layout algorithms~\cite{jin2020visual}~\cite{xie2020visual} are usually proposed to relax visual clutter and hyperedges intersections.
Informative visual metaphor are also designed to illustrate complex hyperedges,
Therefore, the presentation of hypergraphs is difficult to understand and requires a certain learning curve.
Simplifying a complex hypergraph to an easy-to-understand variant node-link graph is a straightforward presentation.
Subset-based hypergraph drawing approaches~\cite{alsallakh2016state}~\cite{qu2021automatic} is designed to address overlap issues of hyperedges.
Hyperedges are illustrated with closed curves enveloping their vertices.
Matrix-based hypergraph drawing~\cite{fischer2020visual} methods can intuitively illustrate the evolution of relationships between sets and elements
with specific matrix layout algorithm, such as Upset~\cite{lex2014upset}.
Parallel-based hypergraph drawing approaches~\cite{valdivia2019analyzing} depicts vertices as equally spaced parallel horizontal lines, and edges as vertical lines.
This kind of hypergraph drawings can be able to address the visual clutter issues.
Compared with matrix-based hypergraphs, subset-based and traditional node-link graph drawing approaches have poor scalability and easily result in visual occlusion and visual clutter.
These hypergraph drawing algorithms aim to illustrate the transaction of graphs, 
and do not consider the direction and the weight of correlations among entities.
The aggregation strategies in these approaches are expired for our weighted directed causal relations.

A directed hypergraph~\cite{ausiello2017directed} is an inherent extension of a traditional graph, as well as a hypergraph is a generation of a generic graph.
A directed hypergraph can also be illustrated by adding a visual channel to a hypergraph that is usually visualized as standard networks or bipartite networks.
The directed hyperedges of each combination causes can be bounded to point to an effect entity, as well as the hyperedges of individual cause can directly point to the effect entities~\cite{tran2020directed}.
Volpentesta et al~\cite{volpentesta2008hypernetworks} proposed a polynomial time algorithm for finding the s-hypernetwork in a directed hypergraph.
Although layout algorithms are introduced to improve the presentation performance, this edge bounded approaches is essentially an extension of the ordinary graph by adding a visual channel to encode the direction of correlations.
Thus, low scalability is a key challenge for this edge bounded approach.
Kinds of applications~\cite{gallo1993directed} are implemented to help users effectively and intuitively obtain the correlations among entities in communication~\cite{ren2020weighted}~\cite{sun2018directed}, data mining~\cite{ranshous2017exchange}, security~\cite{wang2020security}, transportation~\cite{liu2011discovering}~\cite{luo2022directed}, 
and semantic networks~\cite{han2021two}.


Our proposed combined causality can be visualized with a directed hypergraph~\cite{ausiello2017directed}~\cite{tran2020directed}.
The directed hyperedges of each combination causes can be bounded to point to a effect entity, the hyperedges of a individual cause can directly point to the effect entity.
and the causality attribute can be encode with nodes' color.
However, the poor scalability cannot support complex combined causal relations.
A parallel-based hypergraph, namely PAOH, has a high scalability to be used for illustrating hyperedges. 
However, PAOH is unavailable for our directed hypergraph because  
their nodes aggregation and sorting strategies are based on undirected hyperedges.
Therefore, 
we design a visual metaphor of ``electrocircuit" to encode and aggregate ``AND" and ``OR" causes for guaranteeing the non-overlapping and non-intersecting on hyperedges.
We also utilize and embedded diverse ordering operations into a parallel-based, directed and weighted hypergraph to illustrate causal relations.
A focus and context technique and smooth interactions are incorporated into our visualization to help users analyze impelling and inhibiting behavior.




\section{Task Analysis and Pipeline}

\subsection{Task Analysis}
This paper retrieves to visually analyze the individual and combined causality from the event sequence dataset.
To help users obtain the multiple causal relations,
we summarize the following significant analysis tasks in domain literatures as follows:
\begin{compactitem}

	\item[T1] \textbf{Detecting impelling and inhibiting behaviors of causality.} 
	Causality reveals the tight relationships among entities and the characteristics of this causal relation are obvious. 
	Thus,
	impelling or inhibiting causal relations are desired to accurately express the features of causality between entities.

	\item[T2] \textbf{Discovering the combined causality.} 
	A cause combination consists of two or more elements, which individually might not be a cause. 
	These elements in cause combinations interact with each other and have causal relations with other entities.
	The causality of these combination sets are desired to be uncovered. ~\cite{ma2016mining}

	\item[T3] \textbf{Trading-off controllability and effectiveness on cause combinations.}
	Computing all possible combined causes is infeasible due to the exponential number of combined causes. Thus, diminishing redundant cause combination entries and retaining valid cause combination entries is necessary.
	~\cite{ma2016mining}

	\item[T4] \textbf{Providing an informative and intuitive visual metaphor.} 	
	The crossing and 
	overlapping of hyperedges may create visual clutter and visual illusion, which can hinder users from exploring and investigating pattern recognition.
	Diminishing visual illusion and relaxing visual clutter are desired, such as providing an informative and pleasing visual metaphor for the combined causality.
	An informative and intuitive visual representation can help enhance users' understanding of multiple causal relations. 

	\item[T5]  \textbf{Highlighting and displaying causality patterns.} 
	Significant causality patterns may be hidden in a complex causality.
	Exploring a complex causality is time-consuming,
	thus, 
	diverse causal patterns must be illustrated and highlighted to help users significantly improve exploration efficiency.
	~\cite{xie2020visual}~\cite{jin2020visual}CausalFlow~\cite{xie2020causalflow}

	\item[T6]  \textbf{Exploring causal relations at a multilevel.} 
	Users may need to involve different levels of concepts, i.e., the overall trend of causality and focused causality, which is a complex scenario. Hence, a multi-level exploration can help users organize their analysis and improve exploration efficiency.

\end{compactitem}
These requirements help us obtain appropriate visual design principles and make informed decisions about our visual design.

\subsection{ Model Pipeline and User Workflow}
We proposed a combined causal discovery algorithm for detecting impelling and inhibiting behaviors shown in Figure~\ref{Figure:pipline}(B),
After extracting useful entities and filtering valid event sequences from the raw dataset, 
Reactive point processes is employed to model Granger causality for inferring the impelling or inhibiting impact among entities in temporal event sequences.
In this process, eliminating and recruiting principles on candidate cause combinations are defined for ensuring the effectiveness and controllability on cause combinations.
Based on above operations, we obtain a weighted and directed hypergraph dataset, which uncovers the impelling and inhibiting behaviors among entities.
To visually analyze causality,
We design an interactive causality analysis system shown in Figure~\ref{Figure:pipline}(C) for supporting the user analysis task (T1-T6) with the system architecture and user workflow featured in Figure~\ref{Figure:pipline}(A). 
The system shown in Figure~\ref{Figure:overview} includes three panels, namely, combined causality panel, propagation panel, and parameters configuration panel.
With this system, a user can start with an overview of a novel directed parallel-based causality visualization. 
Then, causal patterns are presented in directed and parallel-based visualization with various ordering operations and aggregation strategy for highlighting significant impelling and inhibiting impact patterns. 
Flexible interactions, such as filtering causal strength, brushing, and selecting focused causality, are provided to explore causal relations at different levels of causality abstraction.
Next, users are supported to further modify the incorrect causal relations, analyze impelling and inhibiting behavior patterns, refine and track the causal influence and propagations aided by various ordering operations, smooth interactions.
 


%

\section{Causality Model}
We employ Reactive point processes to model Granger causality in event sequences.
Then, the causality model are trained to fit event sequence dataset by optimizing objective function.
We utilize parameters of the trained model to infer causal strength, as well as impelling or inhibiting behaviors of causality between events.
Finally,
cause combination rules are defined to detect combined causes with one-to-one-relationship causal discovery algorithm.



\subsection{ Granger Causality on Reactive Point Processes}
\label{Section: Granger Causality  on Reactive Point Process}
Inspired by causal discovery~\cite{jin2020visual}, which only uncovers impelling impact among events,
this section describes the extended model for revealing impelling and inhibiting impact among events.
%
The causal discovery~\cite{jin2020visual} use Hawkes processes
to establish a model by relating impelling events.
The occurrence probability of event $u$ is inferred from its conditional intensity function:
\begin{equation}
\begin{aligned}
\lambda_{u}(t)=\mu_{u}+\sum_{u^{\prime}=1}^{U} \int_{0}^{t} \phi_{uu^{\prime}}(s) 
d N_{u^{\prime}}(t-s)     
\end{aligned}
\end{equation}
where the first constant term $\mu_u$ denotes basic entity intensity 
and the second 
term represents the influence of past event on entity $u$ at time $t$.
The value of inferring impact factor is positive, which is a parameter of obtained from the trained Hawkes processes model.
Reactive point processes~\cite{ertekin2015reactive}~\cite{zhang2022optimizing} is an extension of Hawkes processes by relating two types of impelling and inhibiting events.
Hawkes processes only considers the impelling impact of events, while Reactive processes model both impelling and inhibiting effects of chronological events on instantaneous intensity, which can be written as follows:


  \vspace{-0.2cm}

\begin{equation}
\begin{aligned}
\lambda_{u}(t)=\mu_{u} & + \sum_{u^{\prime}=1}^{U} \int_{0}^{t} \phi_{uu^{\prime}}(s)d N_{u^{\prime}}(t-s) \\ 
& - \sum_{u^{\prime}=1}^{U} \int_{0}^{t} \psi_{uu^{\prime}}(s) 
d N_{u^{\prime}}(t-s) 
\end{aligned}
\end{equation}
where $\phi_{uu^{\prime}}$ and $\psi_{uu^{\prime}}$ are kernel functions to reflect the impelling or inhibiting impacts from event $u$ to $u^{\prime}$ respectively.
The intensity function $\lambda_{u}(t)$ combines an inhibiting term for accounting inhibiting effects from past entities and describing the causality intensity among entities. 
Given that $\lambda_u(t)$ describes the causal influence intensity among entities, the value of $\lambda_u(t)$ must be positive to ensure the intensity nonnegativity of events occurrence.
Many researchers utilize a nonlinear function ~\cite{xu2015trailer}~\cite{zhang2022optimizing} to guarantee negativity. Similarly, we leverage a nonlinear function $\hat{\lambda}_{u}(t)$ to guarantee the negativity of $\lambda_u(t)$.

\begin{equation}
\begin{aligned}
\hat{\lambda}_{u}(t)=s \log \left(1+e^{ \frac{\lambda_{u}(t)}{s}}\right)
\end{aligned}
\end{equation}
where $s$ is a small positive constant number, which ensures that we can utilize the property function $g(x) = s\log(1 + exp(x/s)) \approx max\lbrace0,x\rbrace$.

Then, we handle impact function as a linear combination of basis kernel functions, written as follows:

\begin{equation}
\begin{aligned}
\phi_{u u^{\prime}}(t)=\sum_{m=1}^{M} a_{u u^{\prime}}^{m} \kappa_{m}(t) \\
\psi_{u u^{\prime}}(t)=\sum_{m=1}^{M} b_{u u^{\prime}}^{m} \kappa_{m}(t) \\
\end{aligned}
\end{equation}
where $a_{u u^{\prime}}^{m}$ and $b_{u u^{\prime}}^{m}$ are the coefficient of $\kappa_m(t)$, and $a_{u u^{\prime}}^{m} \in[0,1]$, $b_{u u^{\prime}}^{m} \in[-1,0]$. 
Using multiple basis functions improves the accuracy of model estimation but increases computing complexity. 
Thus, a commonly used basic function~\cite{xu2016learning} is employed to balance accuracy and complexity of computation.
To estimate and handle the parameters in our analysis model, we minimize this following objective function:

\begin{equation}
\begin{aligned}
\min \quad -L(\mu, a, b)+\alpha(\|A\|_{F}^{2}+\|B\|_{F}^2)+\beta(\|A\|_{1}+\|B\|_1)
\end{aligned}
\label{equ_obj}
\end{equation}
where $L(\mu, a, b)$ is the Log-Likelihood function, which can be written as follows:


%

\begin{equation}
\begin{aligned}
L(\mu, a, b)=\sum_{c=1}^{C}\left\{\sum_{i=1}^{N_{c}} \log \hat{\lambda}_{u_{i}^{c}}\left(t_{i}^{c}\right)-\sum_{u=1}^{U} \int_{0}^{T_{c}} \hat{\lambda}_{u}(s) d s\right\}
\end{aligned}
\end{equation}
where $\mu$ and $a$ are parameters that must be computed. $\alpha$ and $\beta$ are hyperparameters that control the regularization terms. 
The objective function is a convex function~\cite{zhang2022optimizing}. Thus, we choose the Gradient Descent method to minimize Equ~\ref{equ_obj}. Due to the complexity of integral computation, we employ the Monte Carlo method to compute the integral as follows
(see more details in supplementary materials):

%
\begin{equation}
\begin{aligned}
\int_{0}^{T_{c}} \hat{\lambda}_{u}(s) d s=\frac{T_{c}}{N} \sum_{i=1}^{N} \hat{\lambda}_{u}\left(t^{(i)}\right)  \quad t^{(i)} \sim U\left(0, T_{c}\right)
\end{aligned}
\end{equation}

After obtaining parameters $\mu$, $a$, and $b$, the causal strength of entities $u$ on $u^{\prime}$can be directly inferred from the impact coefficient $a$ and $b$. 
Voting and trading-off strategies are utilized to determine causal strenth by counting the number and the average of $a$ and $b$ in the coefficient.
If entity $u$ has an impelling impact on entity $u^{\prime}$, 
we statistic the average of positive coefficients in basis function from entity $u$ to entity $u^{\prime}$. 
For example,  $M_i$ represent coefficient sets whose basis function coefficients are positive, negative, and zero. 
$M_{max}$ denotes the set with the largest number of elements among $M_i$. 
The causal strength can be inferred as follows:

\begin{equation}
\begin{aligned}
e_{uv}=\frac{1}{\left|M_{max}\right|}\sum_{ \ m \in M_{max}}(a_{uv}^m+b_{uv}^m) \quad \quad
M_{max}=\mathop{\arg\max}\limits_{M_i}{\left|M_i\right|}
\end{aligned}
\end{equation}
Therefore, we obtain a directed and one-to-one-relationship causality graph $G(V,\ E)$, which unveils impelling and inhibiting behaviors on causality. 
Edges are weighted by the causal strength $e_{uv}$,
where positive and negative value of $e_{uv}$ indicate impelling and inhibiting behaviors, respectively. 
Elements in nodes set $V$ are the entities, the edge set $E$ denote the causality between two entities.
Then we leverage above causal discovery algorithm, which reveals impelling or inhibiting behavior of causality, to determine the causality of the following candidate cause combinations on effect entity. 

\subsection{ Cause Combination Discovery}
\label{cause combine}
A cause combination consists of two or more entities, which individually might not be a cause. 
The elements in a cause combination interact with each other and jointly have a causal relation with other entity.
The causality of these combinations are desired to be uncovered.

However, 
computing all possible combined causes is infeasible because the number of cause combination sets is exponential. 
We define ``$a \ \rightarrow \ b$'' a individual causality, where $a$ and $b$ refer to a individual cause and effect entity, respectively.
A combined causality ``$\left\{a, \  c\right\} \ \rightarrow \ b$'' is defined, where $\left\{a, \  c\right\} $ refers to a cause combination and $b$ is a effect entity.
Given the limited entities, the number of possible combined causes is $2^n$, where $n$ is the number of all entities. 
To reduce the number of cause combination, the invalid cause combinations are decreased and useful cause combinations are recruited based on a series of rules. 


\begin{figure}[h]
	\centering
	\vspace{-0.1cm}
	\includegraphics[width=\linewidth]{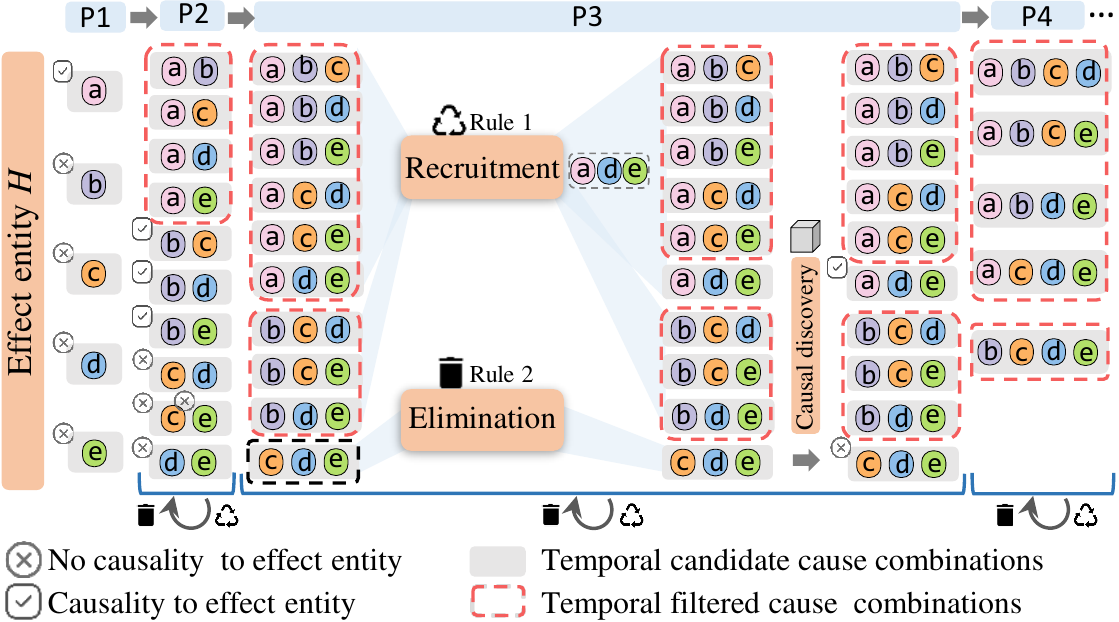}
	\vspace{-0.6cm}
	\caption{
		Illustration of cause combinations process by eliminating the candidate cause combinations and recruiting the filtered combinations with the similarity and co-occurrence of entities.
	}
	\label{Figure:CombinationDiminishing}
	\vspace{-0.5cm}
\end{figure}

To guarantee the effectiveness and controllability of candidate cause combinations (T2, T3),
the following principles are defined.
First,
Determining the maximum number of cause combinations is an efficient way and should be evaluated case by case to decrease the number of candidate cause combinations.
The maximum number of cause combinations determine the number of iterations.
Second, 
to further decrease the number of cause combination,
cause combination of $\left\{a, \  c\right\} $ is regarded as a redundant combination if \textit{c} have been detected as a cause entity.
For example, 
Figure~\ref{Figure:CombinationDiminishing} illustrates cause combination process for a certain effect entity $H$.
If $a$ in P1 is a cause entity of $b$, we do not consider any cause combination entries that containing $a$ in next iteration.
If $\left\{b, \  c\right\} $ in P2 is a cause combination of $H$, we do not consider any cause combination entries containing $ \left\{b, \  c\right\} $.
In this way, the number of candidate cause combination is decreased.

To further improve the effectiveness and usability of candidate cause combinations, we eliminated the invalid cause combinations and recruited the filtered cause combinations.
The similarities of entities are acquired by computing the vectors, which are obtained by training temporal event sequence data with Word2Vec~\cite{mikolov2013efficient}.
The co-occurrence number of entities are calculated from event sequence dataset.
Then,
we leverage similarity and chronology co-occurrence of entity for identifying the possibility of cause combination 
to determine to eliminate or recruit cause combinations. 
Take procedure P3 of Figure~\ref{Figure:CombinationDiminishing} as an example,
elimination or recruitment rules are defined.

\textbf{Eliminating temporal candidate combinations.} 
Candidate cause combinations are obtained shown in the black dotted box after procedure P2,
where certain cause combinations may be redundant.
Thus, these candidate cause combinations must be further filtered (T3) to decrease computation complexity and improve the effectiveness of cause combination. 
Considering that similar entities may together facilitate the occurrence of effect entity,
we utilize Word2Vec to filter the combinations of irrelevant causes.
Given that entities with a low co-occurrence are less likely to be a combination,
we calculate the co-occurrence number of entities to filter cause combinations, in which elements have a low co-occurrence.

\textbf{Recruiting temporal filtered combinations.}
Redundant cause combinations are filtered shown in the red dotted box after procedure P2,
where certain filtered cause combinations may be useful. 
Therefore, these filtered cause combinations must be further confirmed (T3).
Considering that the similar entities may together facilitate the occurrence of the effect entity,
we utilize Word2Vec and co-occurrence to detect cause combinations of similar causes.
Given that entities with a high co-occurrence tend to be a cause combination,
the filtered cause combinations are recruited when the entities in a cause combination have high similarity and high number of co-occurrence.


\begin{algorithm}
	\caption{Cause Combination Algorithm}
	\label{alg1}
	\begin{algorithmic}
		\State \textbf{Input:} Causal Graph $G(V, E)$, Maximum Combination Number $N$
		\State \textbf{Output:} Hyperedges $\Phi$  
		\State Initialize $\Phi = \emptyset$ and candidate cause combinations $\Psi = \emptyset$
		\State Compute candidate cause combinations $\Psi$ with principles
		\For {$i \leftarrow 2$ to $N$}
		\ForAll {$i$-combination $\{v_1, v_2, \dots, v_i\}$}
		\If {$\{v_1, v_2, \dots, v_i\}$ in $\Psi$ and satisfy recruitment rule}
		\State Filter $\{v_1, v_2, \dots, v_i\}$ from $\Psi$
		\EndIf 
		\State Combine $\{v_1, v_2, \dots, v_i\}$ as $v'$
		\State $V' \leftarrow \{v'\} \cup V\setminus\{v_1, v_2, \dots, v_i\}$
		\State Use $V'$ to iterate new causal graph $G'(V',E')$
		\For {cause $u$ in $V'\setminus \lbrace v' \rbrace$}
		\If {$v' \rightarrow u$ in $E'$}
		\State Add $v' \rightarrow u$ to $\Phi$
		\EndIf
		\If {$v'$ not in $\Psi$ and satisfy elimination rule}
		\State Add $v'$ to $\Psi$
		\EndIf
		\EndFor
		\EndFor
		\EndFor
	\end{algorithmic}
\end{algorithm}


Algorithm 1 summarizes the process of candidate cause combinations 
by eliminating the temporal candidate cause combinations and recruiting the temporal filtered cause combinations.
The co-occurrence of entities in a candidate cause combination 
is regarded as a occurrence of a tied entity and fed into our causal discovery algorithm to detect causality.
The impelling or inhibiting impact and causal strength of tied entity to effect entity are determined with causal discovery algorithm described in Section~\ref{Section: Granger Causality  on Reactive Point Process}.

After above principles on candidate cause combination and causal discovery, we obtain a combined causal hypergraph $DHG(V,\ DHE)$, where elements in nodes set $V$ are the entities and the directed hyperedges in $DHE$ denotes the combined causality entries.
This directed and weighted hypergraph is fed into
following visualization component.


\section{ Causality Visualization }
With respect to the combined causality analysis tasks,
this section introduces the details of each visual component, visual
metaphor, and supported interactions. 
The user interface, illustrated in Figure~\ref{Figure:overview}, includes three major components supporting interactive visual exploration: 
a combined causality panel (Figure~\ref{Figure:overview}(A)) for showing the impelling and inhibiting behavior pattern and illustrating focused causality at different levels of information abstraction, 
a parameters configuration panel (Figure~\ref{Figure:overview}(B)) for supporting users to efficiently obtain focused information, 
and a propagation panel (Figure~\ref{Figure:overview}(C)) for tracking the influence propagation when given two entities.


\subsection{ Causality  Aggregation and Ordering  }
{\color{ColorVISRevise}
Combined causality graph is that several factors, which consists of a cause, is a set pointing to an effect factor.
Thus,
}
our combine causality is a directed hypergraph, which inevitably exists a large number of shared nodes and hyeredges intersections.
One of our purposes is to illustrate both individual and combined causality.
Thus, non-overlapping and non-intersecting visualizations are necessary and desired to describe each pair of causality.
Using a one-to-one-relationship graph drawing to present our proposed multiple causality  tends to lead a misleading concept that each entity in combinations can cause an effect independently, resulting in severe visual confusion.
We designed a parallel-based, directed and weighted hypergraph, which implements aggregation strategy, ordering operations, and multi-level exploration for analyzing causal relations.

 \begin{figure}[htb]
 	\centering
 	\vspace{-0.3cm}
 	\includegraphics[width=1\linewidth]{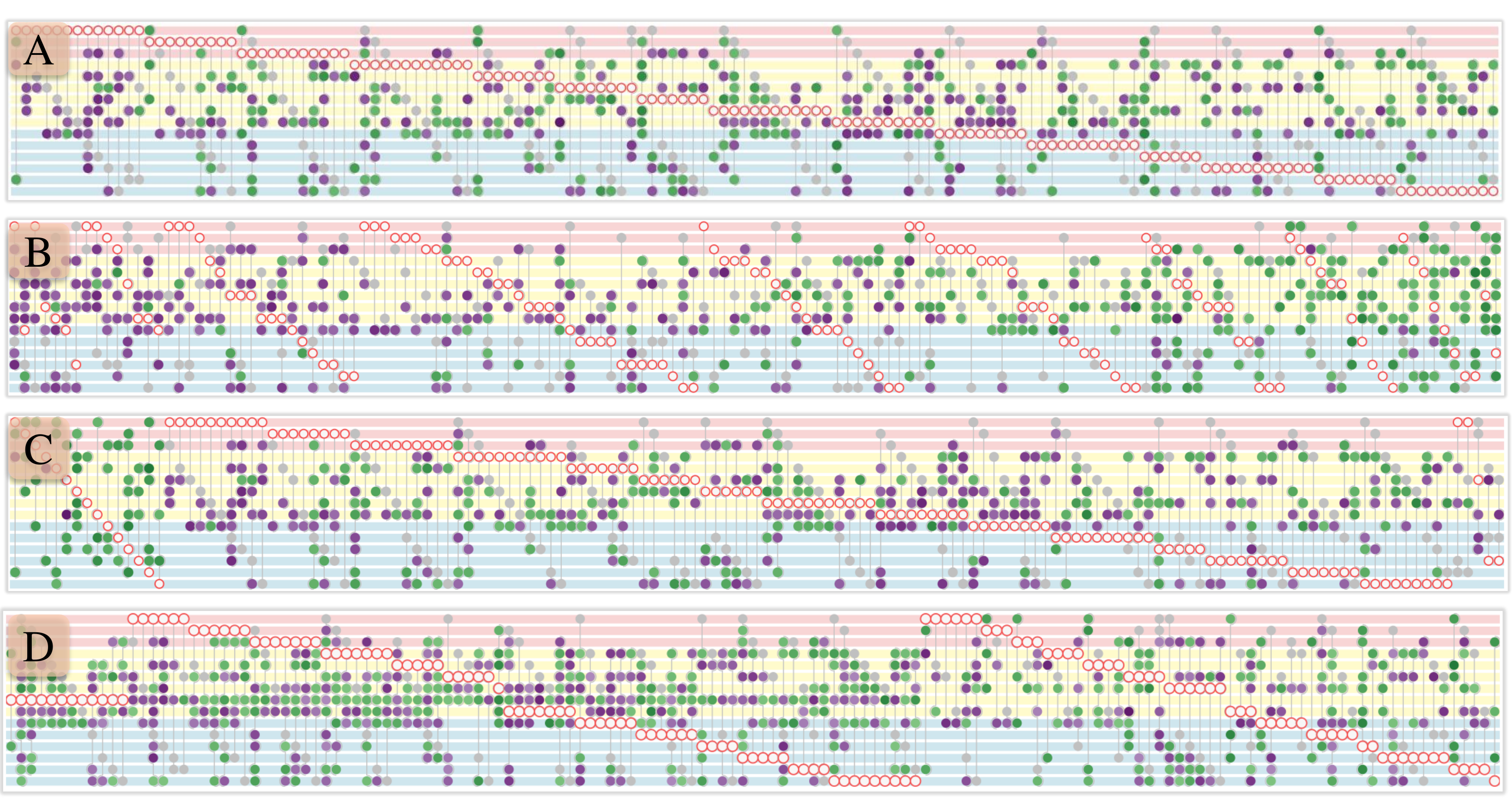}
 	\vspace{-0.6cm}
 	\caption{   
 		Three kinds of hyperedges reordering for causal relations. 
 		(A) Sorting by causal direction. 
 		(B) Sorting by causal strength. 
 		(C) Sorting by the number of causes. 
 		(D) Sorting by a causal topology.
 	}
 	\label{Figure:reordering}
 	\vspace{-0.2cm}
 \end{figure}
 
\begin{figure}[htb]
	\centering
	\vspace{-0.2cm}
	\includegraphics[width=1.06\linewidth]{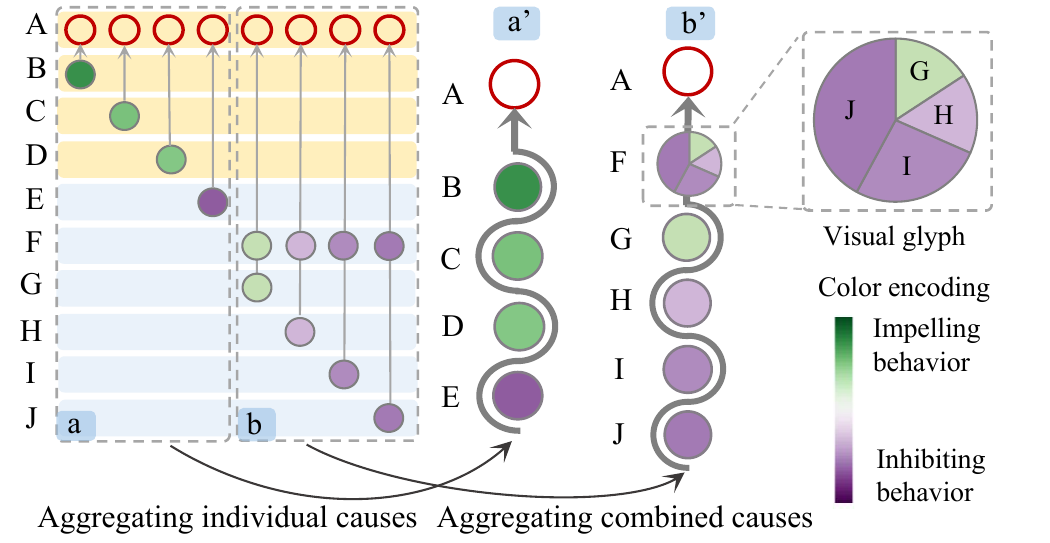}
	\vspace{-0.7cm}
	\caption{   
		Principles of causality aggregation to ensure informativeness and compact of similar cause combinations.
		Left original causality (a)(b) are aggregated to the right metaphors (a')(b').
	}
	\vspace{-0.5cm}
	\label{Figure:MergeCausality}
	
\end{figure}

\textbf{Combined causality visualization (T4).}
We focused on non-overlapping and non-edge-crossing visualizations to describe each pair of combined causality.
We utilize a parallel-based, directed and weighted visualization to illustrate multiple  combined causality.
This causality visualization aims to provide a comprehensive visual abstraction at different levels of causal summarization with a familiar node-link graph to present causal relations.
As illustrated in Figure~\ref{Figure:MergeCausality}(A),
the entities, 
aligned on the left side, are visually encoded with parallel horizontal tracks, 
where the tracks colors describe the communities of this entities, 
solid circles and hollow circles represent starting cause entities and ending effect entities,
a line with an arrow depicts and emphasizes a direction of causality,
and a directed line connecting starting circle pointing to a red ending circle serves as a causality entry.
The opacity of a cause circle represents causal strength.
The color of circles serves as the characteristics of impelling or inhibiting behavior for cause entities, 
where green circle and purple circles serves as the characteristics of the impelling and inhibiting behavior, respectively.


\textbf{Ordering and assignment (T5).}
Sorting and simplifying circles to simplify graphs can facilitate visual aesthetics.
Nodes and hyperedges refer to the entities and combined causes, respectively.
Different orderings on nodes and hyperedges pose various visual clues to help perform diverse tasks, such as bringing similar causality together may uncover communities and bringing top causality may highlight the significant entities.
The layout is modified to various requirements by diverse row ordering and column ordering strategies.

\textit{Row ordering.} Nodes are sorted by ID, groups, alphabetical, and manual interactions, to reveal and highlight the causality patterns, such as central entities, community, or outliers. In our work, following row ordering options are supported:
\begin{compactitem}

	\item
	\textit{Base:} nodes are ordered by effect entities' ID. 
	This strategy aims to move causality belonging to a certain entity together. Users can focused on causality exploration on this entity.

	\item
	\textit{Groups:} nodes are sorted by groups and similarities. We employ Louvain algorithm~\cite{ghosh2018distributed} and Word2Vec for automatic community detection and similarity computation, which highlights potential causality patterns, such as communities and outliers. 
	This ordering tends to reduce edge length when clustering similar entities.

	\item
	\textit{Alphabetical:} nodes are sorted alphabetically. This ordering is beneficial for efficiently searching and locating interested entities by name in long lists.

	\item
	\textit{Manually:} nodes are sorted manually. Sometimes, automatic ordering may not satisfy certain needs, such as reducing the length of the edges. This ordering is designed for manually resorting horizontal entities.		
\end{compactitem}

\textit{Column ordering.} Hyperedges are sorted by ID, causal strength, causal degree, and causal topology to reveal and highlight the causality patterns, such as causal groups and complex causal relations. In our work, users have following vertical ordering options:
\begin{compactitem}

	\item
	\textit{Causal direction:} hyperedges ordered by effect entities' ID are useful in certain application domains shown in Figure~\ref{Figure:reordering}(A)). This ordering moves vertices belonging to a entity close by.

	\item
	\textit{Causal strength:} hyperedges are ordered by causal strength shown in Figure~\ref{Figure:reordering}(B)). This ordering moves stronger causal edges together, which may reveal similar causal edges and enhance significant entities that have strong causal relations.

	\item
	\textit{Degree of cause combination:} hyperedges are ordered by the number of combined causes shown in Figure~\ref{Figure:reordering}(C)).	
	This ordering tends to reveal the overall distribution of individual and multiple causes.

	\item
	\textit{Causal topology:} hyperedges ordered by the direction of a certain entity are useful when focusing on certain interesting entities. This ordering moves related vertices belonging to a causal and effect entity close by and reveals the topology information.

	\item
	\textit{Manually:} hyperedges are ordered manually. This ordering is designed for comparing the similar or interested causality manually.
\end{compactitem}



\begin{figure}[htb]
	\centering
	\vspace{-0.2cm}
	\includegraphics[width=1.02\linewidth]{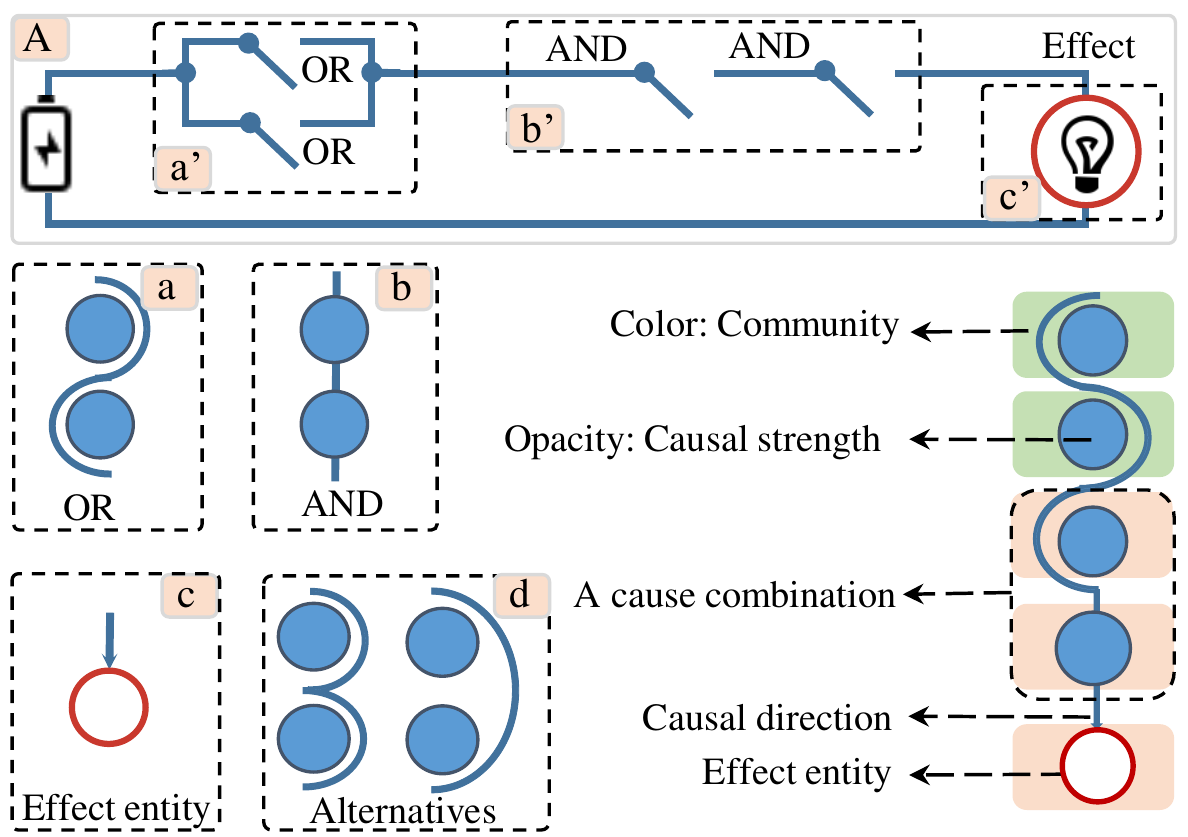}
	\vspace{-0.6cm}
	\caption{   
		Visual metaphors of ``electrocircuit'' (A) for illustrating the aggregated causality relationships. 
		An arc surrounding a circle serves as an optional causal element (``OR''),
		a line crossing a circle encodes the combined causal element (``AND''),  and a red circle encodes a effect entity.
	}
	\label{Figure:VisualMetaphor}
	\vspace{-0.6cm}
\end{figure}

\textbf{Visual metaphor (T4).}
Designing a formative and intuitive visual metaphor for multiple combined causes is an effective and informative method for users to obtain the overall distribution of causal relations.
As illustrated in Figure~\ref{Figure:VisualMetaphor}(A), 
multiple causality are aggregated as an informative visual metaphor of ``electrocircuit'', which is a significant visual component of our visual causal analytics system. 
Operations of merging optional causes (``OR'') and indispensable causes like functions of switches in a parallel circuit and a series circuit.
A closed circuit can light a light bulb.
Causes in Figure~\ref{Figure:VisualMetaphor}(a) and (b) represent ``AND'' or ``OR'' switches in Figure~\ref{Figure:VisualMetaphor}(a') and (b'), respectively. 
An effect entity in the lower left corner of Figure~\ref{Figure:VisualMetaphor}(c) represent a light in Figure~\ref{Figure:VisualMetaphor}(c').
An arc surrounding a circle serves as an optional causal element (``OR''),
a line crossing a circle encodes the combined causal element (``AND''),  and a red circle encodes a effect entity.
The right corner of the Figure~\ref{Figure:VisualMetaphor} illustrates three pairs of causality, where a line crossing a circle is a indispensable cause element (``AND'') and an arc surrounding three circles are optional causes element (OR).
A ``AND'' circle combining with a ``OR'' circle is a cause combination for a effect entity, such as a cause combination in the black dotted area.
In addition, 
two visual metaphors (d) in the lower left corner of Figure~\ref{Figure:VisualMetaphor} are our alternative designs, which tend to lack of pleasing and lead cognition bias on a cause combination.
Thus,
we utilize a pleasing and smoothing curve around optional causes, as well as a direct line crossing indispensable causes.


\textbf{Nodes binding and aggregating (T4, T5).}
Filtering and aggregating vertices to {\color{ColorNameTODO} streamline} graphs are common approaches to ensure readability and informativeness of causal hypergraphs, such as merging hyperedges that share the same entity. 
In our work, 
cause combinations are aggregated as visual metaphors of ``electrocircuit'' to optimize the horizontal space, eliminate demanding context switching and ensure scalability.
As illustrated in Figure~\ref{Figure:MergeCausality}, 
the matrix graph is the original causal hypergraph, where horizontal circles describe the same entity aligned on the left side, the solid and hollow circles encode cause entities and effect entities, respectively.
Solid green or purple circles serve as the combined causes and red hollow circle represents the effect entity.
A line with arrow connecting cause entity and effect entity means a pair of causal relationship.
The visual metaphor of ``electrocircuit'' in the middle of the Figure~\ref{Figure:MergeCausality}(a')(b') are the aggregation results of original causal hypergraph(a)(b).  
We utilize an arc surrounding circles to indicate optional causes (``OR'') in combination and a line crossing circles to encode indispensable causes (``AND'') in combination.
Figure~\ref{Figure:MergeCausality}(a) illustrates four causes, which can independently affect an effect entity respectively.
To save the horizontal space of edges,
four pairs of causal relations are packed and collapsed in figure~\ref{Figure:MergeCausality}(a'), which saves the horizontal space of three edges.
In the same way,
Figure~\ref{Figure:MergeCausality}(b) displays four pairs of causality. 
Entity \textit{F} can be combined with the entity \textit{G}, \textit{H}, \textit{I}, or \textit{J} to affect entity \textit{A}.
These four pairs of causal relations are aggregated in Figure~\ref{Figure:MergeCausality}(b').
In addition,
Pie chart in the upper right corner of Figure~\ref{Figure:MergeCausality}(a) is utilized to illustrate the characteristic of aggregated causality,
The radian size of each sector represents causal strength and 
the color of each sector represent impelling or inhibiting behaviors.



\textbf{Multi-level exploration (T6).}
Depicting causality at a different level of causal abstraction not only provides the distribution of causal information, but also the details of each pair of causality.
Figure~\ref{Figure:overview} (A) shows that
the focused area (a) provides and enhances a user's concerned facts.
Contextual area (c) is always necessary to enhance the expressiveness of a visualization.
Transition area (b) provides a smoothing effect for aggregation/disaggregation from the context area to the focused one.
Controller panel (d) is designed in spired by Multistream~\cite{cuenca2018multistream},
including brushing and linking techniques to help users handle and expand focus or context area. 
The information in a gray area in the controller is zoomed and expanded in the focused area.
The causal relations in the focused area are expanded by dragging the controller. 
When two sides of the controller are dragged, the gray area in the controller is broadened.
When the focused area in the controller is dragged, the length of the contextual area is dynamically updated.
The focused area is broadened by dragging the two sides of the controller. 

\subsection{ Auxiliary Visualization for Causality Exploration}
Confirming and explaining impelling and inhibiting behaviors of causality are necessary, such as the navigation of amending the incorrect causal relations and the influence propagation of the entities.

\textbf{Causality propagation visualization.}
Users always need to drill down into the causality and explore the influence propagation in large searching space.
Causality propagation reveals how the stimulus propagates over and influences entities.
The roots in a causal graph imply sources and the pathways uncover how a source entity propagates over and influences entities.
Causality queries and propagation visualizations are supported in our system by clicking visual encoding of cause and effect entities.
We employ Dijkstra algorithm~\cite{noto2000method} and Stratisfimal layout~\cite{di2021stratisfimal} for identifying and visualizing the shortest influence pathways given the cause and effect entities.





  
\textbf{{\color{ColorNameTODO}Causality reconfirmation visualization.}}
Considering data noise, uncertainty, and complex interaction among entities, our data-driven causal discovery is not always faithful.
The effect entities may be chronologically influenced by a pair of cause entities.
We utilize traditional and easy-understanding multiple histograms to visualize the traffic volume information of each entity.
To some extend,
these multiple histograms can help users confirm certain impelling and inhibiting behaviors of causality and navigate users modifying incorrect causal relations by comparing the chronology of entity occurrence.


\begin{figure*}[htb]
	\centering
	\vspace{-0.3cm}
	\includegraphics[width=1\linewidth]{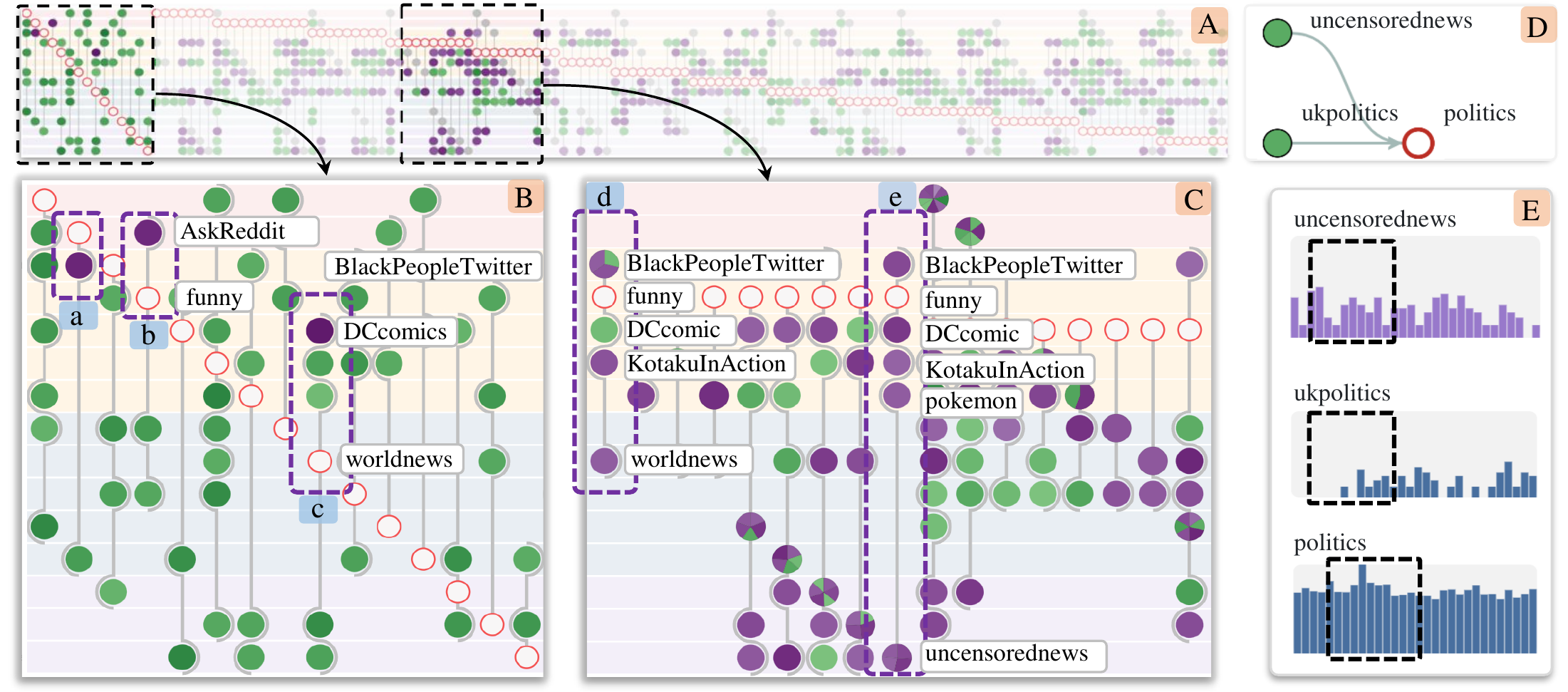}
	\vspace{-0.7cm}
	\caption{   
		(A)~Overall trend of impelling and inhibiting behaviors of subreddits' causality after ascending order by the number of cause combinations. (B)~Focused individual causality and (C)~combined causality after brushing causal relations. (D)~Causal propagations from \textit{ukpolitics} to \textit{politics}. 
	}
	\label{Figure:case1_subreddit}
	\vspace{-0.6cm}
\end{figure*}

\subsection{ User Interactions }
Our work combines various basic and advanced interaction techniques, including multiple views linking, dragging, and brushing, 
to support and enhance the combined causality analysis. 

\textbf{Exploring causality at a multiple levels (T6).} 
Selecting and handling selected causality segments at different levels of abstraction. 
All of the causal relations can be expanded and merged interactively at a regular range number of causal relations by dragging the controller tool and brushing the overview of causal relations.

\textbf{Supporting diverse ordering options(T5).} Entities and hyperedges are sorted by vertical and horizontal ordering approaches ( i.e., grouping similar entities, highlighting causality, and dragging focused entities manually) to efficiently recognize significant causality patterns. 

\textbf{Filtering causality and selecting subgraphs (T5). } The causal strength and the causal combination depth filtering are supported. 
In addition,   
subgraph selection in the causal graph is supported to further explore interested causality.

\textbf{Searching for causal propagations (T6).} Users are supported to search for causal propagations by clicking the arbitrary visual encoding of entities (i. e., circles, layer bands, and texts ) or entering two entities in the input box.

\textbf{Amending spurious causal relations.} 
Considering that automatic causal discovery may not always be reliable, 
users' domain experiences are supported to be incorporated into causal analysis to revise incorrect causality manually. 




%

\section{ Evaluation }
This section evaluates the usefulness and effectiveness of our approach and visual design by using two datasets and implementing a user study.
\vspace{-0.5cm}
\subsection{ Performance of Causality Mechanism}



This section discuss the algorithm performance from following aspects.


\textbf{Time complexity.} For parameter estimation, the time efficiency of our algorithm is \textit{O}(\textit{INC}), where \textit{I} is the iteration number of gradient descents, \textit{N} is the number of entities, and \textit{C} is the number of sequences. 
We examined on a dataset of 9 events and 199 event sequences with Inter Core i5-6500 processor with 8GB memory. 
The Gradient Descent in an individual pair of causality computing takes 50s per iteration. 
The worst time complexity of cause combination is \textit{O}($N^K$), where \textit{K} is the maximum combinations number. 
However, the worst case is rare due to cause combinations filtering rules and parallel computing technology.

\textbf{Space complexity.} For parameter estimation, the space complexity is \textit{O}($N^2M$), where \textit{N} is the number of entities and \textit{M} is the number of basis function. In causality combination, the worst space complexity is \textit{O}($NC_N^K$), where $C_N^K$ is the combination number of entities, but actually the average space complexity is much less than the worst due to the rules filtering in causality combination.

\begin{table}[h]
\centering

\caption{Combination number based on cause combination filtering rules}
\vspace{-0.1cm}
\label{table1}
\begin{tabular}{|l|c|c|c|c|}
\hline
\textbf{Combination Number} & \multicolumn{1}{c|}{\textbf{2}} & \multicolumn{1}{c|}{\textbf{3}} & \multicolumn{1}{c|}{\textbf{4}} & \multicolumn{1}{l|}{\textbf{average value}} 

\\ \hline
\textbf{Original combinations} & 136                                   & 680                                   & 2380                                    & 1065                                 

\\ \hline
\textbf{Filtered combinations}  & 120                                   & 471                                   & 849                                   & 480 

\\ \hline
\textbf{Percentage}  & 11\%                                   & 30\%                                   & 64\%                                   & 55\%
                                     
\\ \hline
\end{tabular}
\vspace{-1.5em}
\end{table}

\vspace{0.1cm}

\textbf{Performance of cause combination.} As mentioned in Section~\ref{cause combine}, we choose several strategies to decrease the search space. When detecting a combined causality, any combinations containing this set are regarded as redundant combinations. 
Therefore, the running time does not reach the worst time complexity in the actual experiment. We test our filter rules in a dataset with seventeen entities. As shown in Table~\ref{table1}. 
In the two-, three-, and four-entity combinations, the filtering rule reduces 11\%, 30\%, and 64\% of combinations, respectively.

\vspace{-0.1cm}
\subsection{ Case Study I: Causality Investigation for Subreddit}
The first one is the subreddit data collected from October to November, 2016.
The dataset covers the top 15 subreddits occurring most frequently after filtering irrelevant entities by Word2Vec and co-occurrence.
A subreddit represents a community with a specific area of interest.
We filter and extract each user’s commenting trajectory, including account name, subreddit and timestamp, on these 15 subreddits, including \textit{News}, \textit{Worldnews}, ... , and \textit{funny}, as an event sequence.


After ranking causality in ascending order by the number of cause combinations, the causality is illustrated in Figure~\ref{Figure:case1_subreddit}(A).
Most nodes in left individual causality are green and most cause combinations are purple. 
Then,
Figure~\ref{Figure:case1_subreddit}(B) illustrates the brushed causality when we focused on individual causality.
Most cause nodes are green, which represents that most of them poses an impelling behavior.
These events are subreddits which receives large attention. 
Popularity is their common feature, therefore similar topics have a mutual incentive causal effect.
After brushing combined causes,
for combined causal relations among subreddits shown in Figure~\ref{Figure:case1_subreddit}(B), most cause nodes are purple, which represent that most of the combined causes belong to inhibiting causal relations.
The inhibiting behaviors easily occur in combined causes.
The capacity of the Internet and the total number of visits are limited due to the limited attention of humans.  
Inhibiting behaviors always exist.
The combination of multiple subreddits receiving large attention does have an inhibiting effect on the attention of other subreddits.

As presented in the purple box of the Figure~\ref{Figure:case1_subreddit}(B), there exists three inhibiting causal relations.
The subreddit \textit{AskReddit} (a) is influenced by \textit{BlackPeopleTwitter},
the subreddit \textit{funny} (b) is influenced by \textit{AskReddit}, 
and the subreddit \textit{worldnews} (c) is influenced by \textit{DCcomics}. 
The subreddit \textit{funny} (d) is a humor depository and 
the subreddit \textit{BlackPeopleTwitter} is a platform of screenshots of Black people being hilarious or insightful on social media.
While \textit{AskReddit} is the place to ask and answer thought-provoking question.
They belong to various types of community,
thus, there exist inhibiting impacts from \textit{BlackPeopleTwitter} to \textit{AskReddit}, and from \textit{AskReddit} to \textit{funny}.
We found \textit{Suicide Squad} was released by \textit{DCcomics} and attracted large attention.
This explosive topic has an inhibiting impact on long-term popular topic \textit{worldnews}.
As shown in the purple box of the Figure~\ref{Figure:case1_subreddit}(C), 
The subreddit \textit{funny} (d), a humor depository, is influenced by four pairs of combinations.
The four cause combinations all contain \textit{BlackPeopleTwitter}.
The two subreddits, \textit{funny} and \textit{BlackPeopleTwitter} belong to humor topic, suggesting that similar subreddits likely cause an inhibiting impact in combined causality.
Similarly,
Figure~\ref{Figure:case1_subreddit}(e) present an inhibiting behavior between   subreddit \textit{KatakuInAction} and \textit{funny}. 

Figure~\ref{Figure:case1_subreddit}(D) shows that when tracking \textit{ukpolitics} and \textit{politics}, a combined causality exists where \textit{ukpolitics} and \textit{uncensorednews} form a cause combination, which has an impelling impact on \textit{politics}.
To confirm the causality, 
we check the attention distributions of subreddits in Figure~\ref{Figure:case1_subreddit}(E),
We find that the histograms of entity \textit{ukpolitics} (e) and \textit{uncensorednews} (f) in a combined cause have a similar peak when \textit{politics} (g) also reaches a peak in corresponding chronological order.
Thus \textit{ukpolitics} and \textit{uncensorednews} have an impelling impact on \textit{politics}.


\vspace{-0.2cm}
\subsection{ Case Study II: Causality Analysis of Website Panels}

The second one comes from Internet Information Server logs for msnbc.com and news-related portions of msn.com for the entire day for September 1999.
Each user's request for a page consists of an event sequence.
There exist 16 entities, including \textit{frontpage}, \textit{news}, \textit{health}, \textit{living}, \textit{weather}, \textit{on-air}, ... , and \textit{business}.

\begin{figure}[htb]
	\centering
	\vspace{-0.4cm}
	\includegraphics[width=1\linewidth]{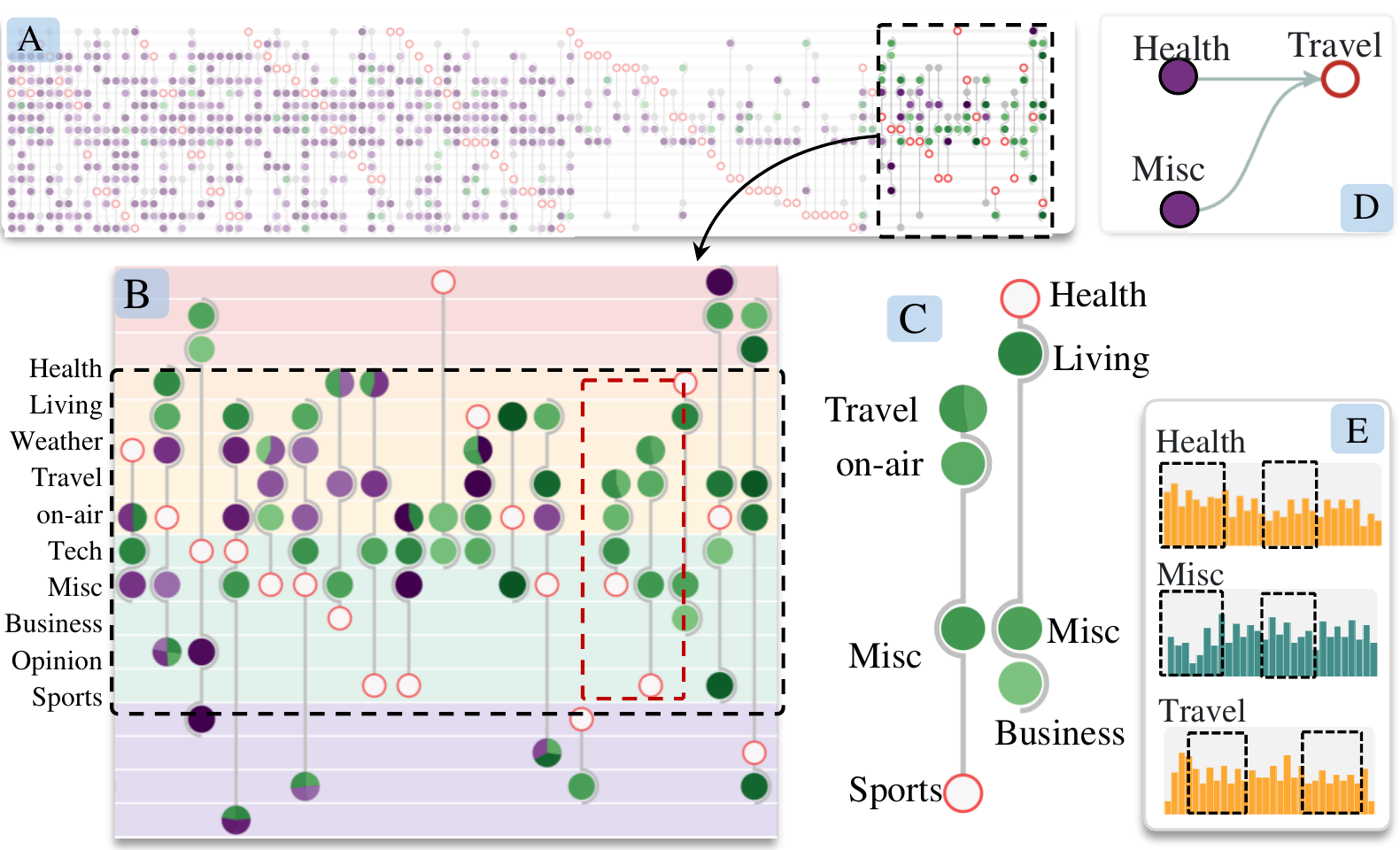}
	\vspace{-0.6cm}
	\caption{   
		(A)~Overall trend of impelling and inhibiting behaviors after sorting by causal strength. 
		(B)~Focused causality after brushing causal relations. 
		(C)~Details of aggregated causality.
		(D)~Target causal propagations. 
		(E)~Distribution of entities' characteristics in chronological order.
	}
	\label{Figure:case2_msnbcdata}
	\vspace{-0.2cm}
\end{figure}

After listing causality in ascending order by causal strengths characteristics, the overall trend of causality shown in Figure~\ref{Figure:case2_msnbcdata}(A), most nodes are purple and a few retained nodes in the black dashed box are green.
This means that most event have an inhibiting behavior and little events have an impelling behavior.

After brushing the green causality, we find these green nodes mainly come from yellow and blue communities shown in Figure~\ref{Figure:case2_msnbcdata}(B).
The entities in the two clusters are daily topics, such as \textit{Health}, \textit{Living}, \textit{Weather}, \textit{Travel}, and \textit{Sports}. 
When hovering over the causality, the details of causal relations are illustrated.
For example, as illustrated in figure~\ref{Figure:case2_msnbcdata}(C), 
two aggregated causality exist. 
The first one is two pairs of combined causality where \textit{Travel} can incorporate \textit{on-air} or \textit{Misc} respectively to influence \textit{Sports}.
The second one is three pairs of causality where \textit{Living}, \textit{Misc}, or \textit{Business} can influence \textit{Health} respectively, suggesting that similar kinds of topics tend to have an impelling behavior on each other.

As shown in Figure~\ref{Figure:case2_msnbcdata}(D), when tracking \textit{Health} and \textit{Travel}, a combined causality exists that \textit{Health} and \textit{Misc} form a cause combination, and purple nodes presents that the cause combination has inhibiting impact on \textit{Travel}.
To confirm the causality, 
we find that the histograms of the former two entities \textit{Health} and \textit{Misc} in the cause combination have two similar periodic peak distributions shown in Figure~\ref{Figure:case2_msnbcdata}(E).
The histograms of the result entity \textit{Travel} have two lead-lag valley values distributions in chronological order.
Thus, we inferred that the combined causes of \textit{Health} and \textit{Misc} have an inhibiting impact on \textit{Travel}.

\vspace{-0.2cm}
\subsection{ User Feedback}

We invited ten participants, who are majored in computer science, to evaluate the effectiveness of the visual encoding of directed combined causality. 
User feedback was also summarized from ten users, who have more than two years of data visualization experience, to gain insights into the effectiveness of our visual encoding and the usefulness and usability of our system. 
We first described the visual encoding and user interactions in our system.
Then system exploration lasted for thirteen minutes for each user to demonstrate the causality patterns that were exposed in case studies. 
User feedback was summarized as the following three aspects:



\textbf{Visual encoding. }
All participants agreed that the visual technique is informative, engaging, and easy to understand when analyzing causality.
They all liked the parallel-based technique because of non-overlapping and non-intersecting characteristics.
They all particularly mentioned that the visual metaphor of ``electrocircuit'' for aggregated causality is impressive. 
According to our participants, user interactions, especially the focus and context technique, are smooth and useful to explore causality from overview to details.
However, 
three of them pointed out that they prefer traditional one-to-one-relationships graphs when analyzing individual causes.

\textbf{Scalability. }
All participants appreciated our combined causality analysis system.
They all confirmed the usefulness and effectiveness of our visual design and interaction techniques.
Seven of them commented that the aggregation strategy and the focus and context technique in our interactive system guarantee the scalability to visualize weighted and directed hypergraph.

\textbf{Improvement.}
The users provided valuable suggestions on how to strengthen our visualization.
Although the system received positive feedback during the interview from the users, they still raised concerns about visual perception about the topology in our visual design.
In addition,
{\color{ColorNameTODO}the system lack of sufficient interpretability and explainability to support decision-making.}
Two of them perceived that more explanations of impelling and inhibiting behaviors should be illustrated for users to better understand causal relations.


\vspace{-0.2cm}
\section{ Discussion}
The experiments on two case studies and a pilot user study confirm the usability and effectiveness of the approach for discovering impelling and inhibiting behaviors, as well as visual causality analysis system.
The present work, however, has some limitations. 

\textit{Time consuming.}
We attempt to integrate user feedback on causal diagnosis into the causal model to update our causality model. 
However,
our causal discovery algorithm is time-consuming. 
Although multiple parallel computations on multiple cores have been used, 
causal discovery can not be computed in real-time.
Thus,
human-in-the-loop can not be reached due to the high complexity of the causal model.

\textit{Visual design.} 
The system provide a {\color{ColorNameTODO} parallel-based and directed hypergraph}, which incorporate a informative visual metaphor of "electrocircuit" to illustrate aggregated causality for guaranteeing our visualization non-overlapping and non-intersecting.
In addition, 
diverse sorting operations, aggregating strategies, and focus-and-context technique are also embedded into our parallel-based, directed and weighted  hypergraph for helping user efficiently analyze complex combined causality.
However,
Compared with a one-to-one-relationship graph,
our parallel-based and directed hypergraph is still complex and the topology information in our ordered and aggregated hypergraph is not easy to be perceived. 
Thus,
more intuitive visualization on combined causality are desired to provide easy-understanding information.


\textit{Pattern identification.}
Our detected causal hypergraph contains significant causal patterns.
Although our system has provided diverse ordering strategies for assisting causal analysis,
experts may need to drill drown to special causality pattern mining.
This work is the first attempt to use parallel-based, directed and aggregated hypergraph to analyze impelling and inhibiting behaviors.
Thus,
causality pattern discovery approaches are still enlarged and designed to automatically identify 
significant impelling and inhibiting behavior patterns.


Future works can consider two improvements, including combined cause discovery and the more intuitive visual encodings for a weighted and directed hypergraph. 
An efficient combined causal discovery is desired to address the high complexity of impelling and inhibiting behavior modeling.
In this way, human verification can be further embedded into the causality algorithm to improve and update causality model in real time.
In addition, 
we attempt to summarize the characteristics of causality patterns and apply image processing technology for extracting causal patterns automatically.

 
\vspace{-0.2cm}
\section{ Conclusion}
In this paper, we utilize Granger causality based on the Reactive point processes for unveiling the inhabiting and impelling behaviors of causality.
We also propose a novel combined causality visualization, which incorporates a causal discovery mechanism to detect combined causes among entities.
The system provides a parallel-based directed hypergraph, which embeds diverse sorting layouts, aggregating strategies, and focus-and-context techniques to help users efficiently analyze complex combined causality.
Furthermore,   
we also conduct two case studies a pilot user study to evaluate the usability and effectiveness of our work.

\bibliographystyle{abbrv-doi}

\bibliography{reference-evosets}

\begin{thebibliography}{10}

\bibitem{alharbi2015conjunctive}
M.~Alharbi and S.~Rajasekaran.
\newblock Conjunctive combined causal rules mining.
\newblock In {\em Proceedings of IEEE International Symposium on Signal
  Processing and Information Technology}, pp. 28--33, 2015.

\bibitem{alharbi2015disjunctive}
M.~Alharbi and S.~Rajasekaran.
\newblock Disjunctive combined causal rules mining.
\newblock In {\em Proceedings of IEEE International Symposium on Signal
  Processing and Information Technology}, pp. 40--45, 2015.

\bibitem{alsallakh2016state}
B.~Alsallakh, L.~Micallef, W.~Aigner, H.~Hauser, S.~Miksch, and P.~Rodgers.
\newblock {The State-of-the-Art of Set Visualization}.
\newblock {\em Computer Graphics Forum}, 35(1):234--260, 2016.

\bibitem{angrist1996identification}
J.~D. Angrist, G.~W. Imbens, and D.~B. Rubin.
\newblock Identification of causal effects using instrumental variables.
\newblock {\em Journal of the American Statistical Association},
  91(434):444--455, 1996.

\bibitem{ausiello2017directed}
G.~Ausiello and L.~Laura.
\newblock {Directed hypergraphs: Introduction and fundamental algorithms—A
  survey}.
\newblock {\em Theoretical Computer Science}, 658:293--306, 2017.

\bibitem{chickering2002optimal}
D.~M. Chickering.
\newblock Optimal structure identification with greedy search.
\newblock {\em Journal of machine learning research}, 3(11):507--554, 2002.

\bibitem{cleveland1984graphical}
W.~S. Cleveland and R.~McGill.
\newblock {Graphical perception: Theory, experimentation, and application to
  the development of graphical methods}.
\newblock {\em Journal of the American Statistical Association},
  79(387):531--554, 1984.

\bibitem{colombo2014order}
D.~Colombo, M.~H. Maathuis, et~al.
\newblock Order-independent constraint-based causal structure learning.
\newblock {\em J. Mach. Learn. Res.}, 15(1):3741--3782, 2014.

\bibitem{cuenca2018multistream}
E.~Cuenca, A.~Sallaberry, F.~Y. Wang, and P.~Poncelet.
\newblock {Multistream: A multiresolution streamgraph approach to explore
  hierarchical time series}.
\newblock {\em IEEE Transactions on Visualization and Computer Graphics},
  24(12):3160--3173, 2018.

\bibitem{deng2021compass}
Z.~Deng, D.~Weng, X.~Xie, J.~Bao, Y.~Zheng, M.~Xu, W.~Chen, and Y.~Wu.
\newblock {Compass: Towards Better Causal Analysis of Urban Time Series}.
\newblock {\em IEEE Transactions on Visualization and Computer Graphics},
  28(1):1051--1061, 2022.

\bibitem{disixSixmethods}
S.~Di~Bartolomeo, A.~Pister, P.~Buono, C.~Plaisant, C.~Dunne, and J.-D. Fekete.
\newblock Six methods for transforming layered hypergraphs to apply layered
  graph layout algorithms.
\newblock {\em Computer Graphics Forum}, 41(3), 2022.

\bibitem{di2021stratisfimal}
S.~Di~Bartolomeo, M.~Riedewald, W.~Gatterbauer, and C.~Dunne.
\newblock {STRATISFIMAL LAYOUT: A modular optimization model for laying out
  layered node-link network visualizations}.
\newblock {\em IEEE Transactions on Visualization and Computer Graphics},
  28(1):324--334, 2021.

\bibitem{ertekin2015reactive}
{\c{S}}.~Ertekin, C.~Rudin, and T.~H. McCormick.
\newblock Reactive point processes: A new approach to predicting power failures
  in underground electrical systems.
\newblock {\em The Annals of Applied Statistics}, 9(1):122--144, 2015.

\bibitem{fischer2020visual}
M.~T. Fischer, D.~Arya, D.~Streeb, D.~Seebacher, D.~A. Keim, and M.~Worring.
\newblock Visual analytics for temporal hypergraph model exploration.
\newblock {\em IEEE Transactions on Visualization and Computer Graphics},
  27(2):550--560, 2020.

\bibitem{fischer2021towards}
M.~T. Fischer, A.~Frings, D.~A. Keim, and D.~Seebacher.
\newblock {Towards a Survey on Static and Dynamic Hypergraph Visualizations}.
\newblock In {\em Proceedings of IEEE Visualization Conference}, pp. 81--85,
  2021.

\bibitem{gallo1993directed}
G.~Gallo, G.~Longo, S.~Pallottino, and S.~Nguyen.
\newblock Directed hypergraphs and applications.
\newblock {\em Discrete applied mathematics}, 42(2-3):177--201, 1993.

\bibitem{ghosh2018distributed}
S.~Ghosh, M.~Halappanavar, A.~Tumeo, A.~Kalyanaraman, H.~Lu,
  D.~Chavarrià-Miranda, A.~Khan, and A.~Gebremedhin.
\newblock {Distributed Louvain Algorithm for Graph Community Detection}.
\newblock In {\em Proceedings of IEEE International Parallel and Distributed
  Processing Symposium}, pp. 885--895, 2018.

\bibitem{guo2020survey}
R.~Guo, L.~Cheng, J.~Li, P.~R. Hahn, and H.~Liu.
\newblock {A survey of learning causality with data: Problems and methods}.
\newblock {\em ACM Computing Surveys}, 53(4):1--37, 2020.

\bibitem{han2021two}
J.~Han, B.~Cheng, and X.~Wang.
\newblock Two-phase hypergraph based reasoning with dynamic relations for
  multi-hop kbqa.
\newblock In {\em Proceedings of International Conference on International
  Joint Conferences on Artificial Intelligence}, pp. 3615--3621, 2021.

\bibitem{jin2020visual}
Z.~Jin, S.~Guo, N.~Chen, D.~Weiskopf, D.~Gotz, and N.~Cao.
\newblock Visual causality analysis of event sequence data.
\newblock {\em IEEE Transactions on Visualization and Computer Graphics},
  27(2):1343--1352, 2020.

\bibitem{jin2012discovery}
Z.~Jin, J.~Li, L.~Liu, T.~D. Le, B.~Sun, and R.~Wang.
\newblock {Discovery of Causal Rules Using Partial Association}.
\newblock In {\em Proceedings of International Conference on Data Mining}, pp.
  309--318, 2012.

\bibitem{kale2021causal}
A.~Kale, Y.~Wu, and J.~Hullman.
\newblock {Causal Support: Modeling Causal Inferences with Visualizations}.
\newblock {\em IEEE Transactions on Visualization and Computer Graphics},
  28(1):1150--1160, 2022.

\bibitem{kaufmann2008subdivision}
M.~Kaufmann, M.~van Kreveld, and B.~Speckmann.
\newblock {Subdivision Drawings of Hypergraphs}.
\newblock In I.~G. Tollis and M.~Patrignani, eds., {\em Proceedings of
  International Symposium on Graph Drawing}, pp. 396--407, 2009.

\bibitem{kumar2015co}
S.~Kumar.
\newblock {Co-authorship networks: A review of the literature}.
\newblock {\em Aslib Journal of Information Management}, 67(1):55--73, 2015.

\bibitem{lex2014upset}
A.~Lex, N.~Gehlenborg, H.~Strobelt, R.~Vuillemot, and H.~Pfister.
\newblock Upset: visualization of intersecting sets.
\newblock {\em IEEE Transactions on Visualization and Computer Graphics},
  20(12):1983--1992, 2014.

\bibitem{li2013mining}
J.~Li, T.~D. Le, L.~Liu, J.~Liu, Z.~Jin, and B.~Sun.
\newblock Mining causal association rules.
\newblock In {\em Proceedings of IEEE International Conference on Data Mining
  Workshops}, pp. 114--123, 2013.

\bibitem{liu2011discovering}
W.~Liu, Y.~Zheng, S.~Chawla, J.~Yuan, and X.~Xing.
\newblock Discovering spatio-temporal causal interactions in traffic data
  streams.
\newblock In {\em Proceedings of ACM SIGKDD International Conference on
  Knowledge Discovery and Data Mining}, pp. 1010--1018, 2011.

\bibitem{luo2022directed}
X.~Luo, J.~Peng, and J.~Liang.
\newblock Directed hypergraph attention network for traffic forecasting.
\newblock {\em IET Intelligent Transport Systems}, 16(1):85--98, 2022.

\bibitem{ma2016mining}
S.~Ma, J.~Li, L.~Liu, and T.~D. Le.
\newblock Mining combined causes in large data sets.
\newblock {\em Knowledge-Based Systems}, 92:104--111, 2016.

\bibitem{mikolov2013efficient}
T.~Mikolov, K.~Chen, G.~Corrado, and J.~Dean.
\newblock Efficient estimation of word representations in vector space.
\newblock {\em arXiv preprint arXiv:1301.3781}, 2013.

\bibitem{mirza2014analysis}
P.~Mirza and S.~Tonelli.
\newblock An analysis of causality between events and its relation to temporal
  information.
\newblock In {\em Proceedings of International Conference on Computational
  Linguistics: Technical Papers}, pp. 2097--2106, 2014.

\bibitem{nandy2018high}
P.~Nandy, A.~Hauser, and M.~H. Maathuis.
\newblock High-dimensional consistency in score-based and hybrid structure
  learning.
\newblock {\em The Annals of Statistics}, 46(6A):3151--3183, 2018.

\bibitem{noto2000method}
M.~Noto and H.~Sato.
\newblock A method for the shortest path search by extended dijkstra algorithm.
\newblock In {\em Proceedings of IEEE International Conference on Systems, Man
  and Cybernetics}, vol.~3, pp. 2316--2320, 2000.

\bibitem{ouvrard2020hypergraphs}
X.~Ouvrard.
\newblock Hypergraphs: an introduction and review.
\newblock 2020.

\bibitem{qu2021automatic}
B.~Qu, E.~Zhang, and Y.~Zhang.
\newblock {Automatic Polygon Layout for Primal-Dual Visualization of
  Hypergraphs}.
\newblock {\em IEEE Transactions on Visualization and Computer Graphics},
  28(1):633--642, 2021.

\bibitem{ramsey2017million}
J.~Ramsey, M.~Glymour, R.~Sanchez-Romero, and C.~Glymour.
\newblock {A million variables and more: the Fast Greedy Equivalence Search
  algorithm for learning high-dimensional graphical causal models, with an
  application to functional magnetic resonance images}.
\newblock {\em International Journal of Data Science and Analytics},
  3(2):121--129, 2017.

\bibitem{ranshous2017exchange}
S.~Ranshous, C.~A. Joslyn, S.~Kreyling, K.~Nowak, N.~F. Samatova, C.~L. West,
  and S.~Winters.
\newblock Exchange pattern mining in the bitcoin transaction directed
  hypergraph.
\newblock In {\em Proceedings of International Conference on Financial
  Cryptography and Data Security}, pp. 248--263, 2017.

\bibitem{ren2020weighted}
J.~Ren, H.~Zhang, Z.~Du, Y.~Sun, H.~Hu, and X.~Zhu.
\newblock {Weighted-Directed-Hypergraph-Based Spectrum Access for Energy
  Harvesting Cognitive Radio Sensor Network}.
\newblock {\em IEEE Access}, 8:68570--68579, 2020.

\bibitem{seth2007granger}
A.~Seth.
\newblock {G}ranger causality.
\newblock {\em Scholarpedia}, 2(7):1667, 2007.

\bibitem{spirtes1991probability}
P.~Spirtes, C.~Glymour, and R.~Scheines.
\newblock From probability to causality.
\newblock {\em Philosophical Studies}, 64(1):1--36, 1991.

\bibitem{spirtes2000causation}
P.~Spirtes, C.~N. Glymour, R.~Scheines, and D.~Heckerman.
\newblock {\em Causation, prediction, and search}.
\newblock 2000.

\bibitem{sun2018directed}
Y.~Sun, Z.~Du, Y.~Xu, Y.~Zhang, L.~Jia, and A.~Anpalagan.
\newblock {Directed-hypergraph-based channel allocation for ultradense cloud
  D2D communications with asymmetric interference}.
\newblock {\em IEEE Transactions on Vehicular Technology}, 67(8):7712--7718,
  2018.

\bibitem{tran2020directed}
L.~H. Tran and L.~H. Tran.
\newblock Directed hypergraph neural network.
\newblock {\em arXiv preprint arXiv:2008.03626}, 2020.

\bibitem{tsamardinos2006max}
I.~Tsamardinos, L.~E. Brown, and C.~F. Aliferis.
\newblock {The max-min hill-climbing Bayesian network structure learning
  algorithm}.
\newblock {\em Machine learning}, 65(1):31--78, 2006.

\bibitem{valdivia2019analyzing}
P.~Valdivia, P.~Buono, C.~Plaisant, N.~Dufournaud, and J.-D. Fekete.
\newblock Analyzing dynamic hypergraphs with parallel aggregated ordered
  hypergraph visualization.
\newblock {\em IEEE Transactions on Visualization and Computer Graphics},
  27(1):1--13, 2019.

\bibitem{vehlow2015state}
C.~Vehlow, F.~Beck, and D.~Weiskopf.
\newblock {The State of the Art in Visualizing Group Structures in Graphs}.
\newblock In {\em Proceedings of Eurographics Conference on Visualization}, pp.
  21--40, 2015.

\bibitem{volpentesta2008hypernetworks}
A.~P. Volpentesta.
\newblock Hypernetworks in a directed hypergraph.
\newblock {\em European Journal of Operational Research}, 188(2):390--405,
  2008.

\bibitem{wang2020security}
B.~Wang, Y.~Sun, T.~Q. Duong, L.~D. Nguyen, and N.~Zhao.
\newblock {Security enhanced content sharing in social IoT: A directed
  hypergraph-based learning scheme}.
\newblock {\em IEEE Transactions on Vehicular Technology}, 69(4):4412--4425,
  2020.

\bibitem{wang2015visual}
J.~Wang and K.~Mueller.
\newblock {The visual causality analyst: An interactive interface for causal
  reasoning}.
\newblock {\em IEEE Transactions on Visualization and Computer Graphics},
  22(1):230--239, 2015.

\bibitem{xie2020visual}
X.~Xie, F.~Du, and Y.~Wu.
\newblock {A Visual Analytics Approach for Exploratory Causal Analysis:
  Exploration, Validation, and Applications}.
\newblock {\em IEEE Transactions on Visualization and Computer Graphics},
  27(2):1448--1458, 2020.

\bibitem{xie2020causalflow}
X.~Xie, M.~He, and Y.~Wu.
\newblock {CausalFlow: Visual Analytics of Causality in Event Sequences}.
\newblock 2020.

\bibitem{xiong2019illusion}
C.~Xiong, J.~Shapiro, J.~Hullman, and S.~Franconeri.
\newblock Illusion of causality in visualized data.
\newblock {\em IEEE Transactions on Visualization and Computer Graphics},
  26(1):853--862, 2019.

\bibitem{xu2016learning}
H.~Xu, M.~Farajtabar, and H.~Zha.
\newblock Learning granger causality for hawkes processes.
\newblock In {\em Proceedings of the International Conference on Machine
  Learning}, vol.~48, pp. 1717--1726. PMLR, 2016.

\bibitem{xu2015trailer}
H.~Xu, Y.~Zhen, and H.~Zha.
\newblock Trailer generation via a point process-based visual attractiveness
  model.
\newblock In {\em Proceedings of the International Conference on Artificial
  Intelligence}, p. 2198–2204. AAAI Press, 2015.

\bibitem{yen2019exploratory}
C.-H.~E. Yen, A.~Parameswaran, and W.-T. Fu.
\newblock {An Exploratory User Study of Visual Causality Analysis}.
\newblock {\em Computer Graphics Forum}, 2019.

\bibitem{zhang2021combined}
H.~Zhang, C.~Yan, S.~Zhou, J.~Guan, and J.~Zhang.
\newblock {Combined cause inference: Definition, model and performance}.
\newblock {\em Information Sciences}, 574:431--443, 2021.

\bibitem{zhang2022optimizing}
X.~Zhang, Y.~Zhou, D.~Wu, M.~Hu, X.~Zheng, M.~Chen, and S.~Guo.
\newblock {Optimizing Video Caching at the Edge: A Hybrid Multi-Point Process
  Approach}.
\newblock {\em IEEE Transactions on Parallel and Distributed Systems}, 2022.
\newblock To be appeared.

\end{thebibliography}
\end{document}